\documentclass[10pt, a4paper]{article}

\usepackage[final]{lrec2026} 

\usepackage{inconsolata}

\usepackage{graphicx}
\usepackage{amsmath}
\usepackage{booktabs}
\usepackage{multirow} 
\usepackage{fancyvrb}

\usepackage{longtable,booktabs}
\usepackage{caption,placeins,dblfloatfix}
\usepackage{booktabs}

\usepackage{tabularx}
\usepackage{multirow}
\usepackage{xcolor}
\usepackage[most]{tcolorbox} 

\usepackage[dvipsnames]{xcolor}
\usepackage{hyperref}

\usepackage{algorithm,algpseudocode}

\usepackage{rotating}   

\newcommand{\highlight}[2]{\colorbox{#1}{#2}}

\definecolor{colorhex1}{HTML}{f9f4e6} 
\definecolor{colorhex2}{HTML}{f2fbf6} 
\definecolor{colorhex3}{HTML}{e6f1f9}

\definecolor{colorhex1_v2}{HTML}{e9d390} 
\definecolor{colorhex2_v2}{HTML}{c1ecd4} 
\definecolor{colorhex3_v2}{HTML}{94c3e5}

\definecolor{CustomBlue}{RGB}{57,83,191}

\usepackage{amssymb}
\usepackage{arydshln}
\makeatletter
\def\adl@drawiv#1#2#3{%
        \hskip.5\tabcolsep
        \xleaders#3{#2.5\@tempdimb #1{1}#2.5\@tempdimb}%
                #2\z@ plus1fil minus1fil\relax
        \hskip.5\tabcolsep}
\newcommand{\cdashlinelr}[1]{%
  \noalign{\vskip 1.3pt
           \global\let\@dashdrawstore\adl@draw
           \global\let\adl@draw\adl@drawiv}
  \cdashline{#1}[.4pt/2pt]
  \noalign{\global\let\adl@draw\@dashdrawstore
           \vskip 3pt}}
\makeatother

\newcommand{\coloredsquare}[1]{\textcolor{#1}{$\blacksquare$}}

\definecolor{custom1}{RGB}{15,157,88}

\newcommand{\wmtplusplus}{\textsc{WMT24++}}

\newcommand{\bleu}{\textsc{BLEU}}
\newcommand{\comet}{\textsc{COMET}}

\newcommand{\salamandra}{\textsc{Salamandra}}

\newcommand{\salamandraseven}{\textsc{Salamandra}-7B}
\newcommand{\salamandratwo}{\textsc{Salamandra}-2B}

\newcommand{\cometkiwi}{\textsc{COMET-}\textsc{\scriptsize Kiwi}}

\newcommand{\gptmini}{\textsc{GPT-4.1-mini}}
\newcommand{\gptnano}{\textsc{GPT-4.1-nano}}

\newcommand{\geminitwo}{\textsc{Gemini-2.0-Flash}}
\newcommand{\geminitwofive}{\textsc{Gemini-2.5-Flash}}
\newcommand{\llamathree}{\textsc{Llama3-8b}}
\newcommand{\gemmathree}{\textsc{Gemma3-27b}}

\newcommand{\madlad}{\textsc{MadLad400-7b}}

\newcommand{\acadata}{\textsc{ACAData}}
\newcommand{\acadtrain}{\textsc{ACAD-train}}
\newcommand{\acadbench}{\textsc{ACAD-bench}}

\title{\textsc{ACAData}: Parallel Dataset of Academic Data for Machine Translation}

\name{%
  {\large\bfseries Iñaki Lacunza\textsuperscript{*}\textsuperscript{1}, Javier Garcia Gilabert\textsuperscript{*}\textsuperscript{1}, Francesca De Luca Fornaciari\textsuperscript{*}\textsuperscript{1},}\\[0.6ex]%
  {\large\bfseries Javier Aula-Blasco\textsuperscript{1}, Aitor Gonzalez-Agirre\textsuperscript{1}, Maite Melero\textsuperscript{1}, Marta Villegas\textsuperscript{1}}%
}

\address{%
  \textsuperscript{*} Core Contributors\\
  \textsuperscript{1} Barcelona Supercomputing Center (BSC)\\
  \texttt{inaki.lacunza@bsc.es, javier.garcia1@bsc.es, fdelucaf@bsc.es}%
}

\abstract{
We present \acadata, a high-quality parallel dataset for academic translation, that consists of two subsets: \acadtrain, which contains approximately 1.5 million human-generated paragraph pairs across 12 languages, and \acadbench, a curated evaluation set of almost 6,000 translations covering 12 directions. To validate its usefulness, we fine-tune two Large Language Models (LLMs) on \acadtrain\ and benchmark them on \acadbench\ against specialized machine-translation systems, general-purpose, open-weight LLMs, and several large-scale proprietary models. Experimental results demonstrate that fine-tuning on \acadtrain\ leads to improvements in academic translation quality by +6.1 and +12.4 d-\textsc{BLEU} points on average for 7B and 2B models respectively, while also improving long-context translation in a general domain by up to 24.9\% when translating out of English. The fine-tuned top-performing model surpasses the best propietary and open-weight models on academic translation domain. By releasing \acadtrain, \acadbench\ and the fine-tuned models, we provide the community with a valuable resource to advance research in academic domain and long-context translation.
 \\ \newline  }

\begin{document}

\maketitleabstract

\section{Introduction}


While English has been long established as the lingua franca for scientific research, a significant volume of impactful work is being published in other languages \cite{repec:spr:scient:v:118:y:2019:i:2:d:10.1007_s11192-018-2987-0}. As a result, Machine Translation (MT) has become essential for extending access to and integrating academic findings. However, building MT systems tailored to the academic domain presents challenges that differ from general-purpose translation (e.g., domain-specific terminology, emerging neologisms, complex syntactic constructions), often leading to reduced translation quality \cite{roussis-etal-2024-enhancing}.

Training state-of-the-art MT systems for less-represented languages and specialized domains requires large amounts of high-quality parallel data. In fact, when training translation systems, the quality of parallel corpora plays a critical role not only during the different stages of model training (e.g. pre-training, supervised fine-tuning, preference optimization, etc.) but also in evaluation. Yet, constructing such corpora is far from easy.


\begin{figure}[t]
\centering
  \includegraphics[width=\linewidth]{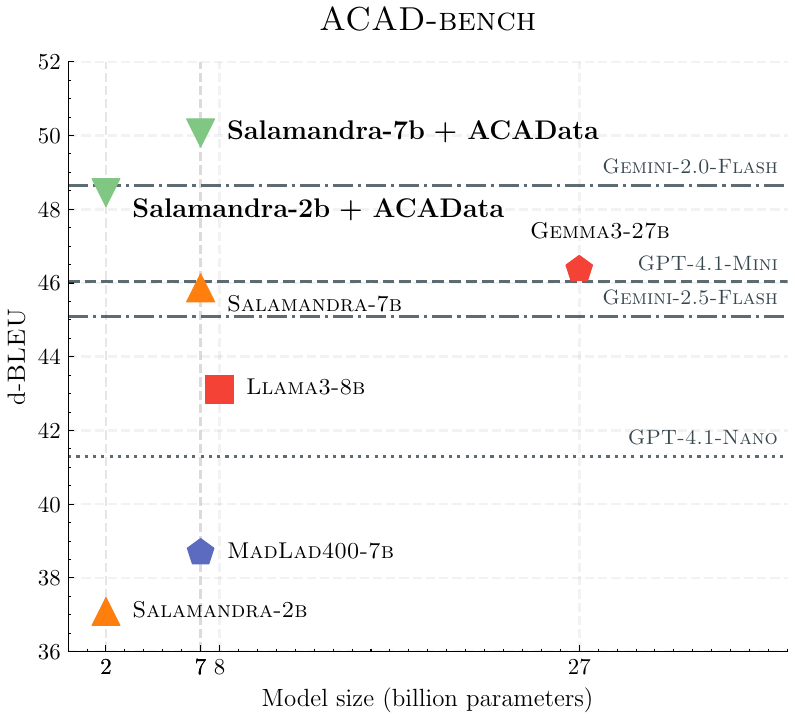}
  \caption{ Translation quality on \acadbench\ in xx→en directions for models fine-tuned with \acadtrain{} and a set of open-weight and proprietary systems of different scales. When the scale is not known, we represent it with a horizontal line. }\label{fig:model_size_dbleu}
\end{figure}

Inspired by prior work leveraging parallel text from public academic repositories, we introduce the \acadata\ dataset: a multilingual parallel corpus extracted from academic abstracts made openly accessible by various research institutions. The dataset contains paragraph-level translations for a wide set of European language pairs in the academic domain, covering specialized content across a broad range of disciplines. All texts are collected from the metadata accompanying the published works and consist of firsthand translations provided by their authors. \acadata\ is divided into two subsets\footnote{\url{https://huggingface.co/datasets/BSC-LT/ACAData}}

\begin{itemize}

    \item \textbf{\acadtrain:} a high-quality parallel training dataset of 1,461,418 paragraph translations across 12 European languages. This human-generated, academic domain-specific data has been obtained from multiple research institutions.

    \item \textbf{\acadbench:} an evaluation set of 5,944 translation instances for benchmarking MT systems on formal, technical, and scientific domains. This test set covers the most frequent language pairs present in the training corpus.
    
    
    
\end{itemize}

We demonstrate the usefulness of the \acadtrain\ dataset by fine-tuning two LLMs and evaluating their performance on \acadbench\ before and after fine-tuning\footnote{\url{https://huggingface.co/BSC-LT/salamandraTA-2B-academic }}\footnote{\url{https://huggingface.co/BSC-LT/salamandraTA-7B-academic}}, as well as in the standard general-domain \wmtplusplus\ dataset. Additionally, we benchmark a range of open-weight and proprietary systems on \acadbench\ to assess their academic specific translation capabilities and provide some reference results on our benchmarking split. Despite their relatively small size, the fine-tuned models achieve a level of translation quality that is on a par with, or even surpasses, that of proprietary models on \acadbench\ (see Figure~\ref{fig:model_size_dbleu}). 

The entire \acadata\ dataset statistics are summarized in Table \ref{tab:dataset_summary}. By open-sourcing our dataset and fine-tuned models, we encourage the machine translation community to pursue research on academic domain translation.


\begin{table}[!ht]
  \centering
  \footnotesize
  \setlength{\tabcolsep}{6pt} 
  \begin{tabular}{lrr}
    \toprule
     & \textbf{\acadtrain} & \textbf{\acadbench} \\
    \midrule
    Instances             & 1,461,418 &   5,944    \\
    Languages             & 12 & 5 \\
    Directions               & 96         &      12     \\
    Src len ($\mu$ ± $\sigma$)    & 1,051 ± 759 & 1,091 ± 779  \\
    Tgt len ($\mu$ ± $\sigma$)    & 1,107 ± 814 & 1,169 ± 832   \\
    \bottomrule
  \end{tabular}
  \caption{Summary statistics for the \acadtrain\ training set and the \acadbench\ benchmark, including number of instances, number of translation directions, and mean (± standard deviation) source (Src) and target (Tgt) paragraph lengths computed using length in characters.}
  \label{tab:dataset_summary}
\end{table}

\section{Related work}

In the field of Machine Translation, large-scale multilingual corpora, often composed of synthetic or web-mined data, predominate. One of the most widely used repositories for MT is OPUS \citep{tiedemann-2012-parallel}, which aggregates nearly all publicly available parallel resources and provides a comprehensive overview of available parallel data. Its collection includes massive multilingual corpora such as CCMatrix \citep{schwenk2020ccmatrixminingbillionshighquality}, NLLB \citep{nllbteam2022languageleftbehindscaling}, HPLT \citep{de-gibert-etal-2024-new}, or NTEU \citep{garcia-martinez-etal-2021-neural}.

As pointed out by \citet{Kreutzer_2022} and \citet{ranathunga2024qualitydoesmatterdetailed}, this web-mined data is generally characterized by issues such as poor alignment, reduced accuracy, high levels of noise, language and domain mismatch, and it typically requires extensive filtering and preprocessing before being suitable for training MT models \cite{steingrimsson-etal-2023-filtering}.

In addition to generic corpora, a number of domain-specific datasets have been developed to support the fine-tuning of MT models on specialized domains. These resources are typically carefully filtered and curated, resulting in higher-quality data. Within the scientific domain, several examples exist. The Scielo corpus \citep{neves-etal-2016-scielo} is a parallel dataset of scientific publications for the biomedical domain in three language pairs: English to Spanish, English to French and English to Portuguese. The data was extracted from Scielo\footnote{\url{https://scielo.org}}, a database of open access scientific publications with a focus on developing and emerging countries. In a related effort, \citet{soares-etal-2018-large} compiled a further version of the Scielo dataset by including parallel data for the directions English to Spanish, Portuguese to Spanish, and English to Portuguese.

Further examples are the Asian Scientific Paper Excerpt Corpus ASPEC \citep{nakazawa-etal-2016-aspec}, a large-size parallel corpus of scientific paper abstracts in Japanese to English and Chinese to Japanese, and the CAPES TDC dataset \citep{soares2018parallel}, a parallel corpus of theses and dissertations abstracts in English and Portuguese collected from the Brazilian CAPES (Coordenação de Aperfeiçoamento de Pessoal de Nível Superior)\footnote{\url{https://www.gov.br/capes/pt-br}} website.
ed it, you can disable auto-conversion:

All these resources have a limited scope, since they are focused on a single domain (e.g., biomedicine) or a reduced number of languages.
More recently, \citet{roussis-etal-2022-scipar} introduced SciPar, a multilingual parallel corpus of theses and dissertations abstracts with 9.17 million sentence pairs in 31 language pairs including data extracted from 86 repositories and archives with openly available metadata. 
SciPar provides coverage across a broad range of scientific disciplines and serves as a key reference and inspiration for the present work. However, this prior work does not include any experimental applications or evaluations demonstrating the usefulness of the corpus for the training or assessment of MT systems, and it is limited by a non-commercial license.
Building on this work, we introduce a new high-quality academic parallel corpus sourced from public institutions and demonstrate its practical value for fine-tuning. We include translation directions that were not covered by previous works. Additionally, we create a manually curated test set covering the most frequent language pairs in our training data, designed to support reliable evaluation of MT performance in the academic domain. Importantly, we release both portions of the dataset and the fine-tuned models under a permissive license, providing a broadly accessible new resource and promoting its unrestricted use across the scientific community.





\section{Methodology}
\label{sec:methodology}

This section describes the end‐to‐end procedure used to construct our dataset, from initial harvesting through final preprocessing.

\subsection{Abstract Pair Harvesting}
\label{subsec:abstract_harvesting}


The core of our dataset consists of parallel paragraphs corresponding to abstracts, which are extracted from a variety of scientific, academic, and governmental repositories (most of them coming from Spanish institutions). All data are harvested via the OAI-PMH~\footnote{\href{https://www.openarchives.org/pmh/}{https://www.openarchives.org/pmh/}} (Open Archives Initiative - Protocol for Metadata Harvesting) interface. 
A complete list of the institutions and their repository URLs is provided in Appendix~\ref{app:data_sources}.

During harvesting, we process each metadata record’s \texttt{<description>} field. Whenever a record contains multiple description elements, we use the LaBSE sentence‐transformer model~\citep{feng-etal-2022-language} to embed each element into a shared multilingual latent space and compute pairwise cosine similarities. Since LaBSE is limited to 512-token sequences, we apply the SLIDE sliding window approach~\citep{raunak2023slide} to process longer inputs. Pairs of descriptions are considered valid candidate translations if their cosine similarity is greater than 0.8. 
Further details about the procedure are provided in Appendix~\ref{app:slide}.

To filter out off-target translations, we apply GlotLID~\citep{kargaran-etal-2023-glotlid} for language identification, discarding any pair in which either paragraph has a probability language score below 0.8. Then, we restrict our dataset to the most represented languages in the source repositories: English, Spanish, French, Catalan, Portuguese, German, Italian, Galician, Basque, Dutch, Greek and Asturian.
Finally, we discard any pair in which either paragraph contains fewer than 40 characters, in order to reduce noise from overly brief instances. Following these filtering steps, we obtain 855,874 parallel paragraphs.

\begin{figure}[!ht]
\centering
  \includegraphics[width=\linewidth]{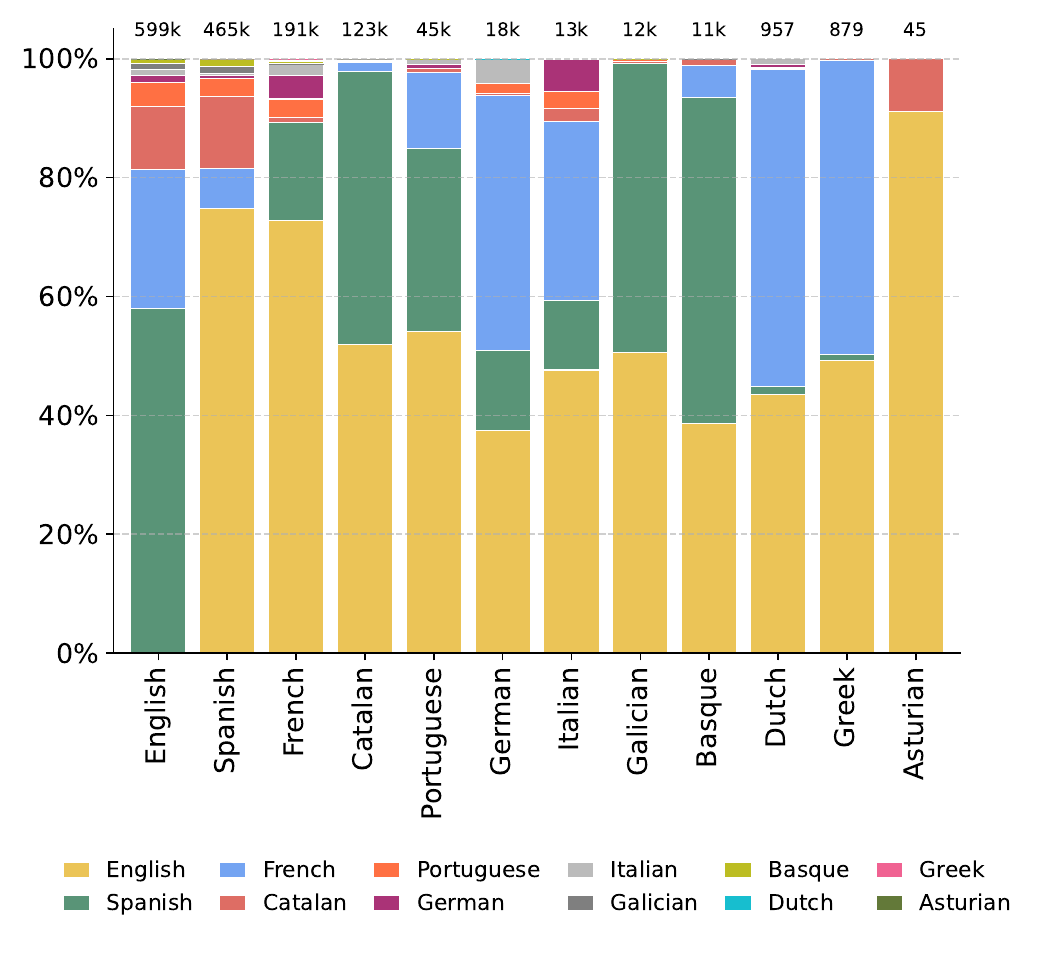}
  \caption{Relative distribution of data across languages in \acadata. Each bar represents a source language, with segments showing the proportion of sentence pairs per target language. Percentages are normalized per language, while absolute pair counts are indicated above each bar.}\label{fig:data_distribution}
\end{figure}

\subsection{Normalization and deduplication}

After collecting all translation pairs, we apply a cleaning and normalization pipeline followed by deduplication to improve consistency and remove redundant entries. 

First, any leading language markers (e.g.,\ “(Spanish)”, “[eng]”) are stripped from each segment. We then normalize punctuation and typography by converting all variants of quotation marks and apostrophes to their ASCII equivalents, replacing masculine ordinals (“º”) with degree symbols (“°”), and converting any superscript or subscript digits to regular digits. Next, we remove common inline markers (short bracketed or parenthesized codes, leading “//” or “:”), collapse simple HTML tags (e.g.,\ <br>, <i>, <b>), and collapse multiple whitespace characters into single spaces.

After normalization, we deduplicate at the pair level by constructing a key from the first 300 characters of each source-target segment and dropping any duplicate keys. This process yields a final set of 733,709 clean, unique translation pairs. We use both directions for each pair, meaning the total number of translation instances is 1,467,418, ready for splitting into training and benchmarking subsets. The relative distribution of translation pairs across languages is shown in Figure \ref{fig:data_distribution}, with absolute counts indicated above each bar. 
An exhaustive analysis is provided in Appendix~\ref{app:dataset_analysis}.

\subsection{Benchmarking Set Splitting \& Curation}
\label{subsec:test_set_splittin-curation}

After constructing the full dataset, we partition it into a large training set and a smaller, manually curated set. The latter contains a total of 2,972 pairs, which we also use in both directions, summing up to 5,944 translation instances. These instances are chosen as follows:

\begin{itemize}
  \item \textbf{Language coverage} The test set includes the six most represented language pairs from the training data (see\ Table~\ref{tab:small_summary_train}).
  \item \textbf{Size distribution} We sample instances from the eligible pairs so that the distribution of language pairs in the test set closely reflects that of the full training set. We use the number of translation pairs in each language pair as a weight to obtain the benchmarking split. 
\end{itemize}

A detailed description of the construction process for \acadbench\ is provided in Appendix~\ref{app:size_distribution}.

After sampling, we manually curate the test set. We use the previously computed cosine similarities as a heuristic to identify possible low-quality translation pairs. Any pair with a similarity score below a threshold of 0.91, is post-edited by native speakers in \acadbench\ covered languages\footnote{In total, 114 instances were post-edited.}.

\section{\acadata}
\label{sec:Dataset}

This section presents the \acadata\ dataset, providing an analysis of its training and benchmarking splits.  
Throughout the tables, we report statistics for only one translation direction per language pair, since cosine similarity scores are symmetrical and source/target character counts simply swap in the reverse direction. 
Various examples illustrating the diversity of language pairs and variation in text length of \acadbench\ can be found in Appendix~\ref{app:examples}.


\subsection{Training split: \acadtrain}

The training set consists of 1,467,418 translation instances. Beyond applying embedding-based similarity filters, language-identification filters, normalization, and deduplication, no further processing is performed. We cover 96 translation directions. Table \ref{tab:small_summary_train} gives an overview of the six most frequent language pairs (accounting for 96.5\% of the \acadata\ training set); all other directions are grouped under “Other”. The last column of the table reports the average cosine similarity between LaBSE embeddings of the paired paragraphs. The relatively high scores reflect the high quality of the translation pairs. 
Full statistics are reported in Appendix~\ref{app:dataset_analysis}.

\begin{table}[!ht]
\centering
\begin{tabular}{l r r r r}
\toprule
\textbf{Lang Pair} & \textbf{Count} & \textbf{Src $\mu$} & \textbf{Tgt $\mu$} & \textbf{Cos $\mu$} \\
\midrule
en–es   & 380,205 & 1,080 & 1,184 & 0.93 \\
en–fr   & 141,972 &  900 &  979 & 0.90 \\
en–ca   &  70,760 & 1,380 & 1,288 & 0.92 \\
ca–es   &  62,264 & 1,176 & 1,204 & 0.98 \\
es–fr   &  32,575 &  915 &  918 & 0.95 \\
en–pt   &  25,916 & 1,010 & 1,042 & 0.90 \\

\cdashlinelr{1-5}
Other   & 25,519 & 1,050 & 1,018 & 0.93 \\
\midrule

\textbf{Overall} & \textbf{739,211} & \textbf{1,070} & \textbf{1,126} & \textbf{0.93} \\
\bottomrule
\end{tabular}
\caption{Summary of the six largest language-pair subsets (all others are aggregated under “Other”) in \acadtrain. Mean paragraph length (measured in characters) and mean cosine similarity. Counts are shown for one direction only (i.e. half of the total bidirectional instances).}
\label{tab:small_summary_train}
\end{table}

\subsection{Benchmarking split: \acadbench}

The benchmarking set comprises the most common language pairs, sampled to preserve a similar size distribution, and contains 5,944 instances in total. Table \ref{tab:small_summary_test} summarizes its statistics. 
Again, a detailed analysis is provided in Appendix~\ref{app:dataset_analysis}.

\begin{table}[!ht]
\centering
\begin{tabular}{l r r r r}
\toprule
\textbf{Lang Pair} & \textbf{Count} & \textbf{Src $\mu$} & \textbf{Tgt $\mu$} & \textbf{Cos $\mu$} \\
\midrule
en–es   & 2,161 & 1,102 & 1,203 & 0.93 \\
en–fr   &   333 &  897 &  969 & 0.91 \\
en-ca   &   210 & 1,190 & 1,115 & 0.92 \\
ca–es   &   188 & 1,290 & 1,317 & 0.98 \\

es–fr   &    46 &  798 &  795 & 0.95 \\

en–pt   &    34 & 1,037 & 1,065 & 0.90 \\

\midrule
\textbf{Overall} & \textbf{2,972} & \textbf{1,091} & \textbf{1,169} & \textbf{0.93} \\
\bottomrule
\end{tabular}
\caption{Summary of \acadbench. Mean paragraph length (measured in characters) and mean cosine similarity. Counts are shown for one direction only (i.e. half of the total bidirectional instances).}
\label{tab:small_summary_test}
\end{table}




To understand the domain coverage of \acadbench, we classify its instances into 26 domains using NVIDIA’s multilingual-domain classifier\footnote{\href{https://huggingface.co/nvidia/multilingual-domain-classifier}{https://huggingface.co/nvidia/multilingual-domain-classifier}}. The 15 most frequent domains account for 94.6\% of the 2,972 instances\footnote{For each translation pair, we employ only one direction, resulting in a total of 2,972 classified instances (5,944 / 2). }. The four largest domains: People and Society (806 instances, $\approx$21\%), Health (376, $\approx$9.8\%), Jobs and Education (325, $\approx$8.5\%), and Science (216,
$\approx$5.7\%), together account for over 45\% of all examples. A second group of domains: Arts and Entertainment (209), Computers and Electronics (189), Books and Literature (149), and Business and Industrial (130), each represent between 3 and 5\% of the benchmark. The remaining 17 domains span a variety of topics: from Law and Government (102) and News (90) down to the smallest categories such as Real Estate (6), Shopping (5), Adult (4), and Online Communities (2), and make up the final $\approx$30\% of instances. This diversity ensures that ACAD-bench covers both widely studied areas and more specialized subjects. 
Appendix~\ref{app:test_dist} provides the list with the 26 classes identified by the classifier, a description of our methodology, and the complete distribution.

\section{Experiments}

To assess the usefulness of the \acadata\ dataset, we fine-tune two LLMs on its training split and evaluate them on the benchmarking split. We then compare their performance against strong baselines. Previous work on adapting LLMs for long-context machine translation typically follow a two-stage training pipeline. In the first stage, models are fine-tuned on sentence-level parallel data. The second stage then adapts the model to handle longer contexts by training it on long-context parallel data \cite{zhang-etal-2018-improving, wu2024adapting}. Following \citet{ramos2025multilingual}, we skip the initial stage and continue supervised fine-tuning on \acadata\ dataset by using two instructed models from the \salamandra\ family of LLMs \citep{salamandra} that have already been trained on sentence-level parallel data in the instruction-tuning stage. These models were pre-trained from scratch on highly multilingual data and then instruction-tuned to improve performance on all the languages covered in \acadata\ dataset. Specifically, we experiment with the instructed versions of \salamandratwo\footnote{\href{https://huggingface.co/BSC-LT/salamandra-2b-instruct}{https://huggingface.co/BSC-LT/salamandra-2b-instruct}} and \salamandraseven\footnote{\href{https://huggingface.co/BSC-LT/salamandra-7b-instruct}{https://huggingface.co/BSC-LT/salamandra-7b-instruct}} LLMs.

\subsection{Formatting}

We format each instruction using the commonly adopted \texttt{chatml} template \cite{chatml} for instruction tuning. 
An example of a formatted input used during training is shown in Appendix Table~\ref{tab:instruction_template}.

\subsection{Implementation details}


We fine-tune \salamandra{} LLMs using the \texttt{FastChat} \cite{zheng2023judging} and \texttt{DeepSpeed} \cite{deepspeed} frameworks on 32 NVIDIA H100 GPUs. 

We train the models for one epoch with a per-GPU batch size of 1 and 16 gradient‐accumulation steps, resulting in an effective batch size of 512. The learning rate was linearly warmed up over the first 85 steps, reaching a peak of \(1\times10^{-5}\), and then decayed using a cosine schedule. \salamandra\ models were trained on a context length of 8,192 tokens, which was sufficient for our training data. Thus, we kept the original maximum sequence length.\footnote{With this configuration, the 2B model required $\approx 16h$ and the 7B model $\approx 39h$ to complete fine-tuning.} 


\begin{table*}
\centering
\small
\tabcolsep 8.9pt
\scalebox{0.97}{%
\begin{tabular}{l l c c c c c}
\toprule
Direction & Model & d-\textsc{BLEU} & \textsc{BP} & \textsc{Blonde} & \textsc{Comet} & \textsc{Comet-Kiwi} \\
\midrule
xx→en   & {\colorbox{colorhex1}{\makebox[3.5cm][l]{\gptmini\ }}}  & 46.03&\textbf{1.00}&0.60&\textbf{0.84}&\underline{0.77} \\
        & {\colorbox{colorhex1}{\makebox[3.5cm][l]{\gptnano }}}	 & 41.30&0.97&0.55&\textbf{0.84}&\textbf{0.78} \\
        & {\colorbox{colorhex1}{\makebox[3.5cm][l]{\geminitwo }}} & \underline{48.65}&\textbf{1.00}&\underline{0.61}&\textbf{0.84}&\underline{0.77} \\
        & {\colorbox{colorhex1}{\makebox[3.5cm][l]{\geminitwofive }}} & 45.10&0.98&0.58&\textbf{0.84}&\underline{0.77} \\
        & {\colorbox{colorhex2}{\makebox[3.5cm][l]{\llamathree$^\dagger$ }}} & 43.12&0.99&0.56&\underline{0.83}&0.76 \\
        & {\colorbox{colorhex2}{\makebox[3.5cm][l]{\gemmathree$^\dagger$ }}} & 46.37&0.98&0.59&\textbf{0.84}&\underline{0.77} \\
        & {\colorbox{colorhex3}{\makebox[3.5cm][l]{\madlad$^\dagger$ }}} & 38.69&0.86&0.51&0.81&\underline{0.77} \\
\cdashlinelr{2-7}
        & \salamandratwo{}$^\dagger$ & 37.09&0.92&0.52&0.82&0.75 \\
        & \quad + {\small \acadtrain\ } & \underline{48.45}&\textbf{1.00}&\underline{0.61}&\underline{0.83}&0.76 \\
        \cdashlinelr{2-7}
        & \salamandraseven{}$^\dagger$ & 45.87&0.99&0.59&\underline{0.83}&0.76 \\
        & \quad + {\small \acadtrain\ } & \textbf{50.07}&\textbf{1.00}&\textbf{0.62}&\textbf{0.84}&0.76  \\
\cdashlinelr{1-7}

en→xx   & {\colorbox{colorhex1}{\makebox[3.5cm][l]{\gptmini\ }}} & 45.01&\underline{0.99}& - &\underline{0.86}&\textbf{0.82}  \\
        & {\colorbox{colorhex1}{\makebox[3.5cm][l]{\gptnano }}} & 43.78&\textbf{1.00}& - &\underline{0.86}&\textbf{0.82}  \\
        & {\colorbox{colorhex1}{\makebox[3.5cm][l]{\geminitwo }}}	& \underline{48.00}&\underline{0.99}& - &\textbf{0.87}&\textbf{0.82} \\
        & {\colorbox{colorhex1}{\makebox[3.5cm][l]{\geminitwofive }}} & \underline{47.75}&\underline{0.99}& - &\textbf{0.87}&\textbf{0.82} \\
        & {\colorbox{colorhex2}{\makebox[3.5cm][l]{\llamathree$^\dagger$ }}} & 39.87&\underline{0.99}&-&0.85&\underline{0.81} \\
        & {\colorbox{colorhex2}{\makebox[3.5cm][l]{\gemmathree$^\dagger$ }}} & 46.29&\underline{0.99}&-&\underline{0.86}&\textbf{0.82} \\
        & {\colorbox{colorhex3}{\makebox[3.5cm][l]{\madlad$^\dagger$ }}} & 36.08&0.82& - &0.83&0.80 \\
\cdashlinelr{2-7}
        & \salamandratwo{}$^\dagger$ & 32.91&0.90& - &0.83&0.78  \\
        & \quad + {\small \acadtrain\ } & 46.86&0.98& - &\underline{0.86}&\underline{0.81} \\
        \cdashlinelr{2-7}
        & \salamandraseven{}$^\dagger$ & 42.55&0.98& - &\underline{0.86}&\underline{0.81} \\
        & \quad + {\small \acadtrain\ } & \textbf{49.20}&0.98& - &\underline{0.86}&\underline{0.81}  \\ 
\midrule 
xx→es   & {\colorbox{colorhex1}{\makebox[3.5cm][l]{\gptmini\ }}} & 60.60&\underline{0.98}&-&\underline{0.86}&\textbf{0.82} \\
        & {\colorbox{colorhex1}{\makebox[3.5cm][l]{\gptnano }}}	 & 57.88&\textbf{0.99}&-&\underline{0.86}&\textbf{0.82} \\
        & {\colorbox{colorhex1}{\makebox[3.5cm][l]{\geminitwo }}}	 & \underline{62.02}&\underline{0.99}&-&\underline{0.86}&\textbf{0.82}  \\
        & {\colorbox{colorhex1}{\makebox[3.5cm][l]{\geminitwofive }}} & 61.43&\underline{0.98}& - &\textbf{0.87}&\textbf{0.82} \\
        & {\colorbox{colorhex2}{\makebox[3.5cm][l]{\llamathree$^\dagger$ }}} & 55.4&\underline{0.98}&-&\underline{0.86}&\underline{0.81} \\
        & {\colorbox{colorhex2}{\makebox[3.5cm][l]{\gemmathree$^\dagger$ }}} & 60.71&\underline{0.98}&-&\underline{0.86}&\textbf{0.82} \\
        & {\colorbox{colorhex3}{\makebox[3.5cm][l]{\madlad$^\dagger$ }}} & 43.44&0.76& - &0.83&\underline{0.81} \\
\cdashlinelr{2-7}
        & \salamandratwo{}$^\dagger$ & 50.09&0.92& - &0.85&0.80 \\
        & \quad + {\small \acadtrain\ } & \underline{61.97}&\underline{0.98}& - &\underline{0.86}&\textbf{0.82} \\
        \cdashlinelr{2-7}
        & \salamandraseven{}$^\dagger$ & 57.55&\underline{0.98}& - &\underline{0.86}&\textbf{0.82} \\
        & \quad + {\small \acadtrain\ } & \textbf{63.60}&\underline{0.98}&-&\underline{0.86}&\textbf{0.82} \\
\cdashlinelr{1-7}

es→xx   & {\colorbox{colorhex1}{\makebox[3.5cm][l]{\gptmini\ }}} &  54.19&\textbf{0.99}&-&\textbf{0.86}&\textbf{0.81} \\
        & {\colorbox{colorhex1}{\makebox[3.5cm][l]{\gptnano }}} & 51.95&\textbf{0.99}&-&\textbf{0.86}&\textbf{0.81} \\
        & {\colorbox{colorhex1}{\makebox[3.5cm][l]{\geminitwo }}}	 & \underline{60.28}&\textbf{0.99}&-&\textbf{0.86}&\textbf{0.81}  \\
        & {\colorbox{colorhex1}{\makebox[3.5cm][l]{\geminitwofive }}} & 57.61&\textbf{0.99}&-&\textbf{0.86}&\textbf{0.81} \\
        & {\colorbox{colorhex2}{\makebox[3.5cm][l]{\llamathree$^\dagger$ }}} & 52.12&\textbf{0.99}&-&\underline{0.85}&\underline{0.80} \\
        & {\colorbox{colorhex2}{\makebox[3.5cm][l]{\gemmathree$^\dagger$ }}} & 57.31&\textbf{0.99}&-&\textbf{0.86}&\textbf{0.81} \\
        & {\colorbox{colorhex3}{\makebox[3.5cm][l]{\madlad$^\dagger$ }}} & 40.13&0.79& - &0.83&\textbf{0.81} \\
\cdashlinelr{2-7}
        & \salamandratwo{}$^\dagger$ & 47.84&0.94&-&0.84&\underline{0.80}  \\
        & \quad + {\small \acadtrain\ } & \underline{60.09}&\textbf{0.99}&-&\textbf{0.86}&\textbf{0.81} \\
        \cdashlinelr{2-7}
        & \salamandraseven{}$^\dagger$ & 55.65&\underline{0.98}&-&\textbf{0.86}&\underline{0.80} \\
        & \quad + {\small \acadtrain\ } & \textbf{61.61}&\textbf{0.99}&-&\textbf{0.86}&\textbf{0.81}
 \\

\bottomrule
\end{tabular}
}

\caption{\small Aggregated results for the xx$\leftrightarrow$en and xx$\leftrightarrow$es translation directions in \acadbench\ dataset. Baselines are grouped into \colorbox{colorhex3}{\textbf{dedicated MMNMT models}}, \colorbox{colorhex2}{\textbf{medium- to small-sized open-weights models}} and \colorbox{colorhex1}{\textbf{large-scale proprietary general models}}. Models with open weights are marked with $^\dagger$. For every metric, the top-scoring system is shown in bold, while the next two best systems for each direction are underlined. }
\label{tab:results_main}
\end{table*}

\begin{figure*}[!ht]
\centering
  \includegraphics[width=\linewidth]{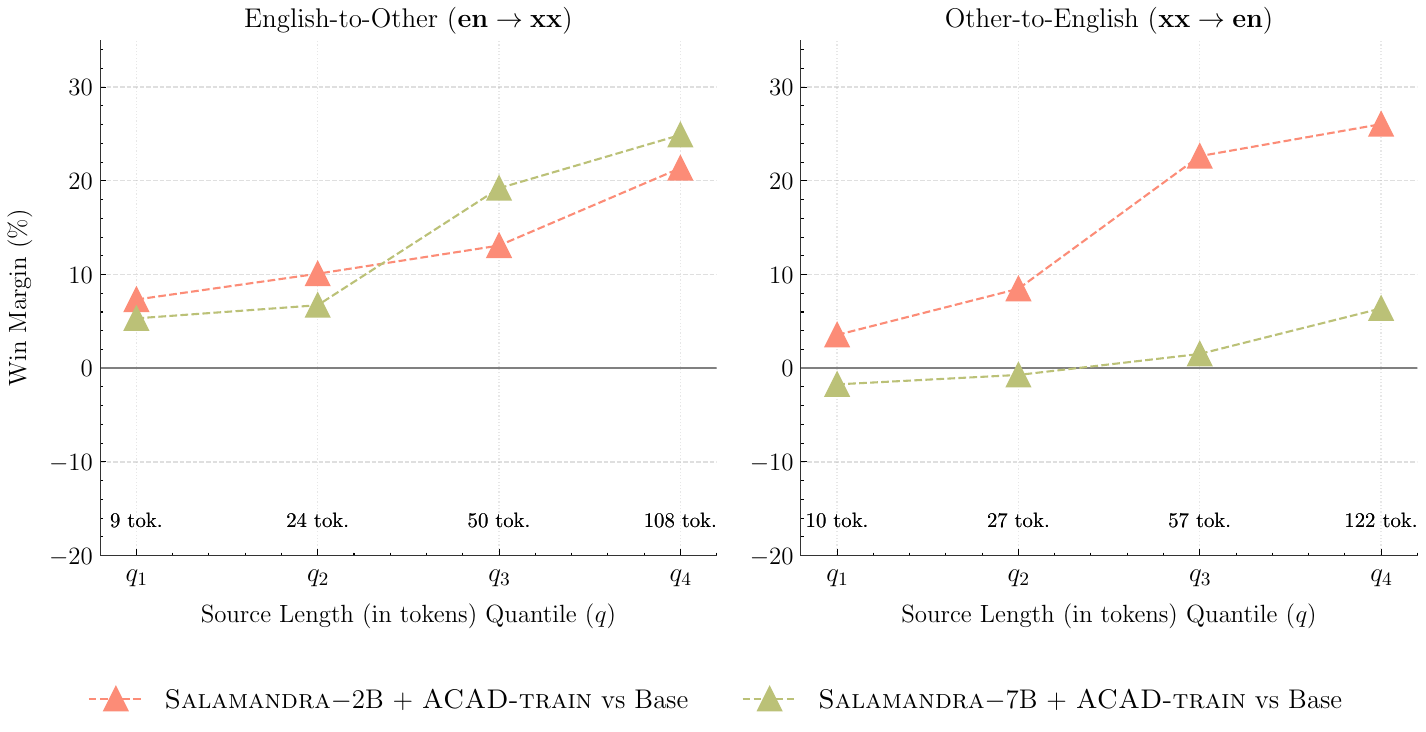}
  \caption{Win margin (\%) over the base model as a function of source length, divided into quartiles (\(q_1\)–\(q_4\)) based on token count. We compare the performance of \salamandratwo\ + {\small \acadtrain} and \salamandraseven\ + {\small \acadtrain} against the base model across translation directions in the \wmtplusplus\ benchmark. The left plot shows results for English-to-Other (en→xx), and the right for Other-to-English (xx→en) directions.}\label{fig:win_margin}
\end{figure*}

\subsection{Evaluation}

\paragraph{Datasets} We conduct experiments on two datasets. Our primary evaluation is on our \acadbench{} dataset, which targets the academic domain. We also use the general-domain \wmtplusplus{} test set \cite{deutsch-etal-2025-wmt24} for two purposes: (i) to assess long-context improvements on the same English-centric language pairs as \acadbench{}, and (ii) to verify that fine-tuning on \acadtrain{} does not degrade performance on general-domain text, using the language directions common to both \acadtrain{} and \wmtplusplus{}.


\paragraph{Baselines} We compare the fine-tuned models with three categories of Machine Translation systems on \acadbench: \coloredsquare{colorhex3_v2} dedicated Massively Multilingual Neural Machine Translation (MMNMT) models,  \coloredsquare{colorhex2_v2} general purpose open-weights LLMs and \coloredsquare{colorhex1_v2} large-scale proprietary LLMs.


\begin{itemize}
    \renewcommand\labelitemi{\coloredsquare{colorhex3_v2}}
    \item \textbf{Dedicated MMNMT models:} we report \madlad\ \cite{10.5555/3666122.3669062}: A widely-used encoder-decoder model supporting more than 400 languages that has been trained with sentence-level parallel data. 

    \renewcommand\labelitemi{\coloredsquare{colorhex2_v2}}
    \item \textbf{General purpose open-weights LLMs:} we use \llamathree~\citep{grattafiori2024llama} and \gemmathree~\citep{team2025gemma} models.

    \renewcommand\labelitemi{\coloredsquare{colorhex1_v2}}
    \item \textbf{Large-scale proprietary LLMs:} we evaluate \gptmini, \gptnano~\citep{achiam2023gpt}, \geminitwo\ and \geminitwofive~\citep{team2023gemini}.
\end{itemize}

\paragraph{Inference}

For the fine-tuned models, \llamathree\ and \madlad\, we perform inference locally using beam search decoding with a beam size of 5. For large-scale proprietary LLMs, we access the models via their respective closed APIs, using their default generation settings. Finally, for evaluating the large-scale open-weight \gemmathree\ model we use Google API. 
More details about inference interfaces can be found in Appendix \ref{app:api_inference}.



\paragraph{Metrics} System performance is evaluated using several metrics targeting different aspects of translation quality. For all translation directions, we report document-level \textsc{BLEU} (d-\textsc{BLEU}\footnote{\texttt{Signature: nrefs:1— case:mixed— eff:no— tok:13a— smooth:exp—version:2.3.1}}) \cite{papineni-etal-2002-bleu} with its brevity penalty (\textsc{BP}).

We additionally evaluate translation quality using two learned regression-based metrics: \comet\footnote{\href{https://huggingface.co/Unbabel/wmt22-comet-da}{https://huggingface.co/Unbabel/wmt22-comet-da}} \cite{rei-etal-2022-comet} and \cometkiwi\footnote{\href{https://huggingface.co/Unbabel/wmt22-cometkiwi-da}{https://huggingface.co/Unbabel/wmt22-cometkiwi-da}} \cite{rei-etal-2022-cometkiwi}. Learned regression-based metrics have been proved useful to evaluate translation quality at the paragraph level \cite{deutsch-etal-2023-training}.

To evaluate discourse-level phenomena, we use \textsc{Blonde} \cite{jiang-etal-2022-blonde}, a metric designed to capture discourse coherence through a set of automatically extracted features. Following \citet{vernikos-etal-2022-embarrassingly}, we only evaluate \textsc{Blonde} when translating into English, since this metric depends on entity taggers and discourse markers trained exclusively in English.

Finally, to evaluate  long-context improvements, we report the win margin based on \textsc{BLEU} on the \wmtplusplus\ dataset for the English-centric directions in \acadbench\ across different source sentence lengths. 
Details on how win margin is computed can be found in Appendix~\ref{app:win_rate}. 

Additionally, we report \bleu\ and \comet\ scores for the fine-tuned models for each of the 62 directions shared between \wmtplusplus{} and \acadtrain{} in Appendix \ref{app:results} and analyze the overall performance in the following section.



\begin{figure*}[!ht]
\centering
  \includegraphics[width=\linewidth]{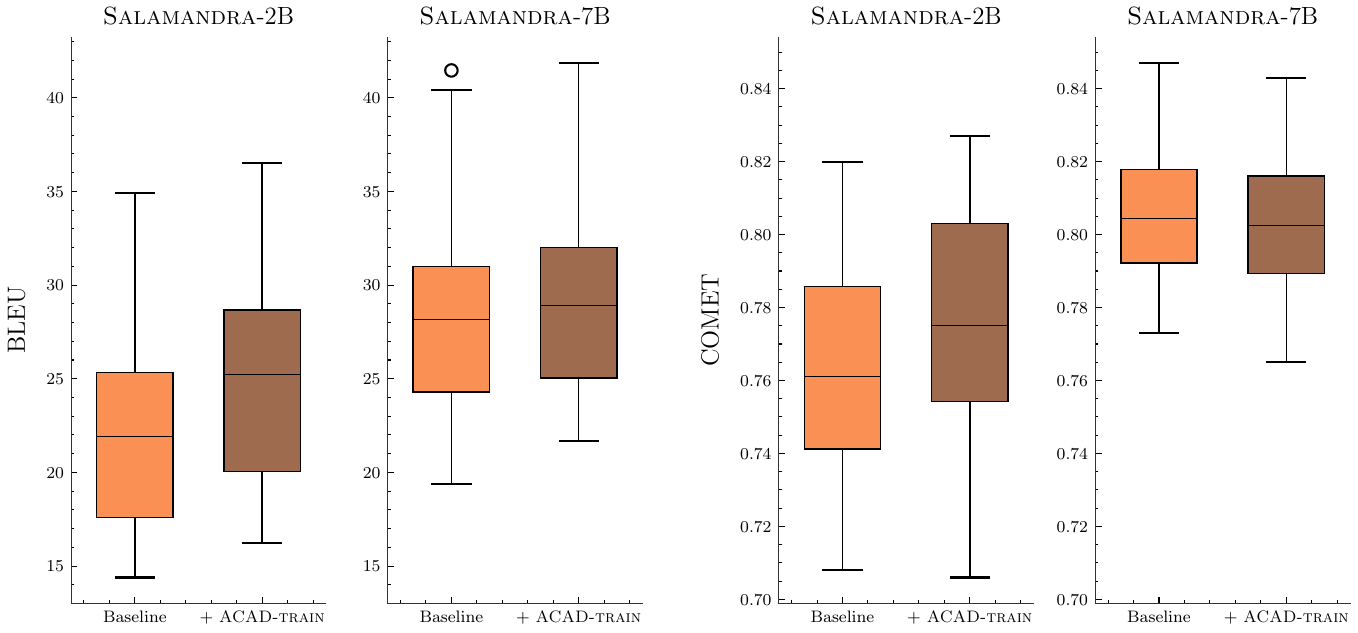}
  \caption{Boxplots for the 62 directions shared between \wmtplusplus\ and \acadtrain. The plots show the distribution of BLEU scores (left) and COMET scores (right). Each boxplot compares the model's performance on \wmtplusplus\ dataset before (Baseline) and after fine-tuning on the \acadtrain\ dataset (\textsc{+} \acadtrain). }\label{fig:box_plots}
\end{figure*}

\subsection{Results}

Table \ref{tab:results_main} presents the performance of all evaluated systems on \acadbench. We aggregate results for all models considering xx→en, en→xx, xx→es and es→xx directions. We find that fine-tuning on \acadtrain\ improves the 7B and 2B fine-tuned models by an average of +6.08 and +12.36 d-\textsc{BLEU} points over the base model, respectively. The fine-tuned models consistently outperform all other evaluated models, including large-scale proprietary systems in d-\textsc{BLEU} metric. We also report \madlad, a MMNMT model in Table  \ref{tab:results_main}. One of the reasons why \madlad\ may be performing worse is that it has been trained on sentence-level corpora, while \acadbench\ evaluates paragraph-level translation. 

Notably, \salamandraseven{} + {\small \acadtrain} is the top-performing model, achieving the highest d-\textsc{BLEU} score across all translation directions. It outperforms the strongest proprietary baseline, \geminitwo{}, by an average of 1.39 d-\textsc{BLEU} points. Even the smaller fine-tuned 2B model outperforms most proprietary models and significantly closes the gap with \geminitwo{}. 

For \textsc{Blonde}, which evaluates discourse and context understanding, the fine-tuned models improve by +0.03 and +0.09 points over the base 7B and 2B models respectively. This result demonstrates a stronger ability to handle context-dependent translation phenomena.  However, on semantic metrics like \comet{} and \cometkiwi{}, the fine-tuned models achieve similar performance as the other baselines, with proprietary models slightly leading in English-centric directions. 


\paragraph{Improved quality in long-context translation} Figure \ref{fig:win_margin} reports win margin on the same English-centric language pairs as \acadbench, evaluated on \wmtplusplus. Translation quality is divided into four quartiles by source sentence length (measured with the model’s tokenizer). The performance gap between the fine-tuned models and their base counterparts grows as the length of the source sentence increases when translating out of English. In fact, when translating out of English, we observe an average win ratio improvement of 24.9\% and 21.3\% in the longest quartile (q4) for the 7B and 2B models respectively. On the contrary, when translating into English, the 2B model shows a strong 26\% improvement in win margin, compared to the 7B model's average gain of 6.4\% in the longest quartile. These results are consistent with the improvements in the brevity penalty (\textsc{BP}) reported for both the 7B and 2B models in Table \ref{tab:results_main}, indicating that fine-tuned models are better able to maintain output length parity in long-context translations.







\paragraph{Improved \bleu\ performance across \wmtplusplus\ language directions} Figure \ref{fig:box_plots} shows the summary of results when fine-tuning with the \acadtrain\ dataset, evaluated on the \wmtplusplus\ test set. The results demonstrate a clear and positive impact on translation quality as measured by \bleu. The 2B model fine-tuning's average score increases by +2.64 \bleu. The larger 7B model also shows improvements, with its average score improving by +1.09 \bleu. In contrast, the impact on \comet\ scores, which measure semantic similarity, is smaller. The 2B model's \comet\ score improves by +0.01 points on average, while the 7B model sees a minor decrease of -0.01.

Overall, these findings indicate that adapting models to the academic domain leads to small improvements on lexical-based metrics, while maintaining comparable performance on semantic-based metrics in a general domain dataset. We hypothesize that fine-tuning with academic data encourages translations that are more technical and formal in style. As a result, improvements are greater on lexical-based metrics such as \bleu, while metrics that focus on semantic similarity, like \comet, show minor changes.  We also find that smaller models tend to benefit more from fine-tuning on \acadtrain, whereas larger models show more stable performance across metrics in a general domain dataset, especially when translating into English.



\section{Conclusions}


In this work, we describe the creation process of the \acadata\ dataset, a novel, high-quality parallel corpus of academic abstracts. Comprising approximately 1.5 million human-generated paragraph pairs (\acadtrain) and a manually curated evaluation set of nearly 6,000 translations (\acadbench), our dataset provides accurate, high-quality translations in the scientific domain. By fine-tuning two open-weight LLMs on \acadtrain, we demonstrate consistent gains in academic translation quality, up to +12.4 d-\textsc{BLEU} for the 2B parameter model and +6.1 d-\textsc{BLEU} for the 7B parameter model, outperforming proprietary models on \acadbench.

Beyond the academic domain, we also show that training with \acadtrain\ yields significant improvements in long-context translation on a standard general-domain benchmark: in the longest source-length quartile, the fine-tuned models achieve up to a 24.9\% gain in win ratio when translating out of English.

By releasing \acadtrain, \acadbench, and the fine-tuned models under permissive licenses, we offer the community a robust foundation training dataset and evaluation benchmark for advancing the development of machine translation systems in the academic domain. 

Ultimately, with this work, we aim to help bridge communication across the global scientific community, and make research more discoverable and accessible regardless of the language it was originally published in.


\section*{Limitations}

In this work, we explain how we built \acadata\ and employed its benchmarking split to evaluate a set of models. However, the scope of the evaluated systems remains limited; future research could extend this analysis to a broader range of models. Additionally, since several baseline systems are proprietary, it is not possible to assess whether the data collected to create \acadata\ was included in their training data, which could potentially influence the results.


\section*{Ethical Statement}

This work focuses on describing the creation of a dataset for academic machine translation sourced from public data, and the fine-tuning of machine translation systems using its training set. The impact of fine-tuning on potential biases, such as gender bias, is left out of the scope of this work. All models and datasets used in our experiments are based on publicly available resources, that may contain inherent biases.

\section*{Acknowledgements}

This work/research has been promoted and financed by the Government of Catalonia through the Aina project.

This work is funded by the Ministerio para la Transformación Digital y de la Función Pública and Plan de Recuperación, Transformación y Resiliencia - Funded by EU – NextGenerationEU within the framework of the project ILENIA with reference 2022/TL22/00215337, 2022/TL22/00215336, 2022/TL22/00215335, 2022/TL22/00215334

This work has been supported and funded by the
Ministerio para la Transformación Digital y de la
Función Pública and the Plan de Recuperación,
Transformación y Resiliencia – funded by the EU
through NextGenerationEU, within the framework
of the Modelos del Lenguaje project.


\nocite{*}
\section*{References}\label{sec:reference}

\bibliographystyle{lrec2026-natbib}
\bibliography{custom}

\appendix



\section{Data Sources}
\label{app:data_sources}

This appendix contains a complete list of the names and URLs of the source repositories from which the translation pairs are obtained, along with the number of pairs extracted from each source before deduplication. This information is shown in Table~\ref{tab:three_column_two_col_sources}.

\begin{table*}[t]
  \centering
  \small
  \scalebox{0.85}{%
  \begin{tabular}{ccc}
    \begin{tabular}[t]{@{}l r@{}}
      \toprule
      \textbf{Repository} & \textbf{Pairs} \\
      \midrule
      
      \href{https://al-qantara.revistas.csic.es/index.php/al-qantara/oai}{al-qantara} & 442 \\
      \href{https://analescervantinos.revistas.csic.es/index.php/analescervantinos/oai}{anales cervantinos} & 279 \\

      \href{https://anuariomusical.revistas.csic.es/index.php/anuariomusical/oai}{anuario musical} & 326 \\
      \href{https://arbor.revistas.csic.es/index.php/arbor/oai}{arbor} & 1265 \\
      
      \href{https://xn--archivoespaoldearte-53b.revistas.csic.es/index.php/aea/oai}{archivo español de arte} & 528 \\
      \href{https://aespa.revistas.csic.es/index.php/aespa/oai}{archivo español de arqueología} & 528 \\
      \href{https://repositorio.aemet.es/oai/request}{arcimis} & 975 \\
      \href{https://rabida.uhu.es/oai/openaire4}{arias montano} & 10150 \\
      \href{https://arqarqt.revistas.csic.es/index.php/arqarqt/oai}{arqueología de la arquitectura} & 280 \\
      \href{https://asclepio.revistas.csic.es/index.php/asclepio/oai}{asclepio} & 791 \\
      \href{https://journal.auditio.com/auditio/oai}{auditio} & 72 \\
      \href{https://bcnroc.ajuntament.barcelona.cat/oai/request}{bcnroc} & 3475 \\
      \href{https://biotecnia.unison.mx/index.php/biotecnia/oai}{biotecnia} & 345 \\
      \href{https://repositorio.uloyola.es/oai/driver}{brújula} & 133 \\
      \href{https://www.riarte.es/oai/driver}{cgate} & 240 \\
      \href{https://clavesjuridicas.com/index.php/raj/oai}{claves jurídicas} & 15 \\
      \href{https://collectaneabotanica.revistas.csic.es/index.php/collectaneabotanica/oai}{collectanea botanica} & 239 \\
      \href{https://dataverse.csuc.cat/oai}{cora} & 101 \\
      \href{https://digital.csic.es/dspace-oai/request}{csic} & 9588 \\
      \href{https://estudiosgallegos.revistas.csic.es/index.php/estudiosgallegos/oai}{cuadernos de estudios gallegos} & 295 \\
      \href{https://cultureandhistory.revistas.csic.es/index.php/cultureandhistory/oai}{culture history d-journal} & 265 \\
      \href{https://dau.url.edu/oai/request}{dau} & 2478 \\
      
      \href{https://ojs.diffundit.com/index.php/index/oai}{diffundit} & 766 \\
      \href{https://digiuv.villanueva.edu/oai/request}{digi uv} & 320 \\
      
      \href{https://digitum.um.es/digitumoai/request}{digitium} & 23268 \\
      \href{https://docta.ucm.es/rest/oai/openaire4}{docta} & 20990 \\
      \href{https://ebuah.uah.es/oai/request}{e-buah} & 9094 \\
      \href{https://edatos.consorciomadrono.es/oai}{e-cienciadatos} & 1028 \\
      \href{http://repositorio.ucjc.edu/oai/driver}{e-ucjc} & 43 \\
      \href{https://ebiltegia.mondragon.edu/oai/driver}{ebiltegia} & 3310 \\
      \href{https://www.euskalmemoriadigitala.eus/oai/request}{emd} & 617 \\
      \href{https://emerita.revistas.csic.es/index.php/emerita/oai}{emerita} & 337 \\
      \href{https://enfermeriacuidandote.com/oai/}{enfermería cuidándote} & 52 \\
      \href{https://estudiosamericanos.revistas.csic.es/index.php/estudiosamericanos/oai}{estudios americanos} & 472 \\
      \href{https://estudiosgeograficos.revistas.csic.es/index.php/estudiosgeograficos/oai}{estudios geográficos} & 726 \\
      \bottomrule
    \end{tabular}
    &
    \begin{tabular}[t]{@{}l r@{}}
      \toprule
      \textbf{Repository} & \textbf{Pairs} \\
      \midrule

      \href{https://estudiosmedievales.revistas.csic.es/index.php/estudiosmedievales/oai}{estudios medievales} & 947 \\
      \href{https://fs.revistas.csic.es/index.php/fs/oai}{fs} & 661 \\
      \href{https://gredos.usal.es/oai/request}{gredos} & 31665 \\
      \href{https://helvia.uco.es/oai/request}{helvia} & 7980 \\
      \href{https://hispania.revistas.csic.es/index.php/hispania/oai}{hispania csic} & 750 \\
      \href{https://cocoon.huma-num.fr/exist/crdo/oai.xql}{humanum} & 1056 \\
      \href{https://www.juntadeandalucia.es/agriculturaypesca/ifapa/oai}{ifapa} & 346 \\
      \href{https://www.larramendi.es/poligrafos_y_autores/en/oai/oai_aut.cmd}{ignacio larramendi} & 566 \\
      \href{https://www.indteca.com/ojs/index.php/Revista_Scientific/oai}{indteca} & 477 \\
      \href{https://repositori.irta.cat/oai/request}{irta} & 25 \\
      \href{https://revista.isciii.es/index.php/bes/oai}{isciii} & 38 \\
      \href{https://rjb.revistas.csic.es/index.php/rjb/oai}{jardín botánico de madrid} & 458 \\
      
      \href{https://maco.csuc.cat/oai/request}{maco} & 1228 \\
      \href{https://manipulus.ucm.es/oai/request}{manipulus} & 1 \\
      \href{https://openaccess.uoc.edu/oai/request}{o2} & 41462 \\
      \href{https://ojs.diffundit.com/index.php/rsa/oai}{ojs} & 766 \\
      \href{https://oai.openedition.org/}{open edition} & 244720 \\
      \href{https://www.uco.es/ucopress/ojs/index.php/psye/oai}{pse} & 87 \\
      \href{https://produccioncientificaluz.org/index.php/rvg/oai}{produccion cientifica luz} & 88 \\
      
      \href{https://pubmed.ncbi.nlm.nih.gov/}{pubmed} & 22813 \\
      \href{https://repositorio.usj.es/rest/oai/request}{r-usj} & 517 \\
      \href{https://centracs.es/revista/oai}{rccs} & 65 \\
      \href{https://revista.cortesgenerales.es/rcg/oai}{rcg} & 409 \\
      \href{https://clavesjuridicas.com/index.php/raj/oai}{rcj} & 15 \\
      \href{https://repositori.tecnocampus.cat/oai/request}{rdtecnocampus} & 59 \\
      \href{https://redined.educacion.gob.es/oai/driver}{redined} & 1250 \\
      \href{https://dspace.umh.es/oai/request}{rediumh} & 6354 \\
      \href{https://sjar.revistas.csic.es/index.php/sjar/oai}{sjar} & 902 \\
      \href{https://tdx.cat/oai/request}{tdx} & 31858 \\
      \href{https://rua.ua.es/server/oai/request}{ua} & 25258 \\
      \href{https://ddd.uab.cat/oai2d}{uab} & 96620 \\
      \href{https://repositorio.uam.es/oai/driver}{uam} & 11066 \\
      \href{https://revistas.uasd.edu.do/index.php/ecos/oai}{uasd} & 217 \\
      \href{https://revistes.ub.edu/index.php/index/oai}{ub} & 19339 \\
      \href{https://riubu.ubu.es/oai/request}{ubu} & 1554 \\
      
      \bottomrule
    \end{tabular}
    &
    \begin{tabular}[t]{@{}l r@{}}
      \toprule
      \textbf{Repository} & \textbf{Pairs} \\
      \midrule
      
      \href{https://rodin.uca.es/oai/driver}{uca} & 4427 \\
      \href{https://repositorio.ucam.edu/oai/driver}{ucam} & 1036 \\
      \href{http://bibliotecas.ucasal.edu.ar/ws/oai2_7}{ucasal} & 1 \\
      \href{https://ruidera.uclm.es/oai/request}{uclm} & 2502 \\
      \href{https://revistas.ucm.es/index.php/oai/oai}{ucm} & 7844 \\
      \href{https://ruc.udc.es/oai/request}{udc} & 24273 \\
      \href{https://udimundus.udima.es/oai/driver}{udimundus} & 43 \\
      \href{https://repositori.udl.cat/oai/driver}{udl} & 1 \\
      \href{https://ddfv.ufv.es/rest/oai/request}{ufv} & 354 \\
      \href{https://digibug.ugr.es/oai/request}{ugr} & 22448 \\
      \href{https://ruja.ujaen.es/oaiextended/request}{uja} & 1057 \\
      \href{http://www.openarchives.org/OAI/2.0/oai_dc/}{uji} & 17223 \\
      \href{https://revpubli.unileon.es/ojs/index.php/index/oai}{ulerevistas} & 1487 \\
      \href{https://accedacris.ulpgc.es/oai/request}{ulpgc} & 7198 \\
      \href{https://revistas.uni.edu.pe/index.php/devenir/oai}{uni peru} & 195 \\
      \href{https://dspace.unia.es/oai/request}{unia} & 633 \\
      \href{https://revistasdigitales.uniboyaca.edu.co/index.php/index/oai}{uniboyaca} & 360 \\
      \href{https://repositorio.unican.es/oai/request}{unican} & 15338 \\
      \href{https://digibuo.uniovi.es/oai/driver}{uniovi} & 246 \\
      \href{https://reunir.unir.net/oai/request}{unir} & 5383 \\
      \href{https://dialnet.unirioja.es/oaites/OAIHandler}{unirioja} & 556 \\
      \href{https://revistas.upc.edu.pe/index.php/docencia/oai}{universidad peruana de ciencias} & 687 \\
      \href{https://zaguan.unizar.es/oai2d}{unizar} & 226 \\
      \href{https://revistas.unla.edu.ar/saludcolectiva/oai}{unla} & 599 \\
      \href{https://upcommons.upc.edu/oai/driver}{upc} & 44275 \\
      \href{https://oai-repositori.upf.edu/oai/request}{upf} & 8269 \\

      \href{https://addi.ehu.es/oai/request}{upv/ehu} & 16360 \\
      \href{https://apps.urv.cat/fedora/oai}{urv} & 1230 \\
      \href{https://revistas.usil.edu.pe/index.php/index/oai}{usil} & 371 \\
      \href{https://roderic.uv.es/oai/request}{uv} & 9212 \\
      \href{https://uvadoc.uva.es/oai/driver}{uva} & 12002 \\
      \href{http://repositori.uvic.cat/oai/request}{uvic} & 3462 \\
      \href{https://zenodo.org/oai2d}{zenodo} & 386 \\
      \href{https://revzoilomarinello.sld.cu/index.php/zmv/oai}{zoilomarinello} & 740 \\
      \\
      \bottomrule
    \end{tabular}
  \end{tabular}
  }
  \caption{Sources of translation pairs with paragraph counts before deduplication.}
  \label{tab:three_column_two_col_sources}
\end{table*}

\section{Handling Large Instances with SLIDE}
\label{app:slide}

The sliding-window adaptation for LaBSE constructs document embeddings by (i) splitting a document into sentences, (ii) forming overlapping, token-bounded windows of whole sentences, (iii) encoding each window with LaBSE, and (iv) aggregating window embeddings into a single L2-normalized document vector using weights that compensate for sentence overlap. This approach allows each sentence to be represented in multiple local contexts while respecting LaBSE's input-size constraint, producing document-level embeddings suitable for cosine-similarity comparisons.

At a high level, the procedure is:
\begin{enumerate}
  \item Split the document \(D\) into sentences \(S=[s_0,\dots,s_{n-1}]\).
  \item Measure token lengths \(t_s\) for each sentence with the model tokenizer.
  \item Greedily pack consecutive sentences into windows so each window's concatenated tokens do not exceed the token budget \(T\).
  \item Slide the window with overlap (windows share sentences), encode each window with LaBSE to obtain embeddings \(e_k\), and record which sentence indices each window covers.
  \item Compute de-overlap weights that apportion each sentence's token mass across the windows that include it, aggregate the weighted window embeddings and L2-normalize to obtain the final document embedding \(E_D\).
\end{enumerate}

\smallskip
\noindent\textbf{Mathematical formulation.} Let \(t_s\) be the token length of sentence \(s\), let \(I_k\) be the set of sentence indices covered by window \(k\), and let \(e_k\in\mathbb{R}^d\) be the embedding of window \(k\). Define the coverage count
\[
c_s \;=\; \#\{k : s\in I_k\},
\]
and the per-window de-overlap weight
\[
w_k \;=\; \sum_{s\in I_k} \frac{t_s}{c_s}.
\]
The unnormalized document embedding is
\[
\widetilde{E}_D \;=\; \sum_k w_k \, e_k,
\]
and the final document embedding is the L2-normalized vector
\[
E_D \;=\; \frac{\widetilde{E}_D}{\lVert \widetilde{E}_D\rVert_2}.
\]
Pairwise similarity between two documents is computed as the dot product \(E_{D_1}^\top E_{D_2}\), equivalent to cosine similarity for unit vectors. The pseudocode is shown in Figure~\ref{alg:sliding_labse_encode}.

\begin{algorithm*}[!ht]
\begin{algorithmic}[1]
  \Procedure{SLIDE\_LaBSE\_Encode}{$D,\,\text{tokenizer},\,\text{LaBSE\_model},\,T,\,B$}

    \If{$D$ is empty}
      \State \Return \(\mathbf{0}\)
    \EndIf

    \State \(S \gets \text{SentenceSplit}(D)\) \Comment{list of sentences \(s_0 \dots s_{n-1}\)}

    \ForAll{sentence \(s\) in \(S\)}
      \State \(t[s] \gets \text{TokenLength}(\text{tokenizer}, s)\)
    \EndFor

    \State \(\text{windows} \gets []\); \(\text{window\_idxs}\gets[]\)
    \State \(i \gets 0\)
    \While{\(i < |S|\)}
      \State \(j \gets i\); \(\text{tok\_sum}\gets 0\)
      \While{\(j < |S|\) \textbf{and} \(\text{tok\_sum} + t[j] \le T\)}
        \State \(\text{tok\_sum} \gets \text{tok\_sum} + t[j]\)
        \State \(j \gets j + 1\)
      \EndWhile
      \If{\(j = i\)} \Comment{single sentence exceeds budget}
        \State \(j \gets i + 1\)
      \EndIf
      \State \(\text{windows.append}(\,\text{JoinSentences}(S[i{:}j])\,)\)
      \State \(\text{window\_idxs.append}(\, [i,\dots,j-1]\,)\)
      \State \(\text{stride} \gets \max(1, \lfloor (j-i)/2 \rfloor)\) \Comment{~50\% overlap}
      \State \(i \gets i + \text{stride}\)
    \EndWhile

    \State \(\text{window\_embs} \gets []\)
    \ForAll{batch in \(\text{Batched}(\text{windows}, B)\)}
      \State \(\text{embs} \gets \text{LaBSE\_model.encode}(\text{batch})\)
      \State append \(\text{embs}\) to \(\text{window\_embs}\)
    \EndFor

    \State \(n \gets |S|\); \(\text{coverage}[0{:}n-1]\gets 0\)
    \ForAll{idxs in \(\text{window\_idxs}\)}
      \ForAll{sentence index \(s\) in idxs}
        \State \(\text{coverage}[s] \gets \text{coverage}[s] + 1\)
      \EndFor
    \EndFor
    \ForAll{\(s = 0\) \textbf{to} \(n-1\)}
      \If{\(\text{coverage}[s] = 0\)} \State \(\text{coverage}[s]\gets 1\) \EndIf
    \EndFor

    \State \(\text{weights}\gets[]\)
    \ForAll{idxs in \(\text{window\_idxs}\)}
      \State \(\text{w}\gets \sum_{s\in \text{idxs}} \frac{t[s]}{\text{coverage}[s]}\)
      \State \(\text{weights.append}(\text{w})\)
    \EndFor

    \State \(\widetilde{E} \gets \mathbf{0}\)
    \For{\(k=1\) \textbf{to} \(|\text{window\_embs}|\)}
      \State \(\widetilde{E} \gets \widetilde{E} + \text{weights}[k] \cdot \text{window\_embs}[k]\)
    \EndFor
    \State \(E_D \gets \widetilde{E} / \lVert \widetilde{E}\rVert_2\)

    \State \Return \(E_D\)
  \EndProcedure
\end{algorithmic}
\caption{High-level pseudocode for SLIDE-style LaBSE document encoding}
\label{alg:sliding_labse_encode}
\end{algorithm*}

The returned vector \(E_D\) is unit-length; document pairs may be compared via the dot product, which equals cosine similarity for normalized vectors.

\section{Dataset Analysis}
\label{app:dataset_analysis}

In this appendix, we present a detailed analysis of the \acadata{} dataset. For each language pair, we report the number of paragraph pairs, length statistics (in characters), and cosine similarity scores between source and target embeddings, which serve as a proxy for semantic relatedness. In Table \ref{tab:lang_code_map}, we show the mapping between the BCP-47 language code and the language name for the languages covered in \acadtrain.

Table~\ref{tab:summary_train} summarizes \acadtrain\ main statistics. All language pairs exhibit strong alignment, with all mean and median cosine similarities above 0.85, and most above 0.90, indicating that the translated paragraphs closely match the source semantics. Average source and target lengths hover around 1,000 characters, while large standard deviations reflect a wide variation in paragraph length, proving the diversity of the dataset.

Table~\ref{tab:summary_test} shows the summary of \acadbench. Every pair exceeds a mean similarity of 0.90 (overall mean ± std; median = 0.93 ± 0.03; 0.93), underscoring the high semantic fidelity of the test translations. As with the training data, paragraph lengths in the test set vary considerably.

For both cases, in order to reduce redundancy, only one direction is shown. Changing the translation direction would imply switching the source length and target length values, while the cosine similarity scores would remain the same.

\section{\acadbench\ Extraction}
\label{app:size_distribution}

We want to make sure that the benchmarking set reflects the same language-pair distribution as the full training data while remaining manageable in size. First, we identify all language pairs (unordered) in \acadtrain\ with at least 1,000 total pairs.  Let \(P\) be this set of eligible language pairs.  We then gather all candidate test-set paragraph pairs whose source–target languages belongs to \(P\).  To sample up to 3,000 test pairs, we draw without replacement in proportion to each pair's frequency in the training set, yielding a preliminary pool of \(3{,}000\) distinct paragraph pairs. Next, we enforce a minimum-per-pair threshold: we discard any language pair for which fewer than 10 sentences have been sampled.  This pruning leaves exactly the six most frequent language pairs from the training data.  Finally, for each selected sentence pair \((s,t)\), we generate two directional translation instances \((s\to t)\) and \((t\to s)\), resulting in a total of \(2 \times 2{,}972 = 5{,}944\) test instances. Algorithm~\ref{alg:testset_sampling} summarizes the followed procedure.

\section{\acadtrain\ Examples}
\label{app:examples}

This appendix contains examples from the \acadtrain\ dataset reflecting its diversity in terms of paragraph length and language coverage. The examples highlight the wide range of translation directions, the variability of paragraph lengths, and the dataset's focus on academic content. The examples are shown in Figures~\ref{fig:example1}-\ref{fig:example7}. In Table \ref{tab:instruction_template}, we additionally show an example of a formatted instruction following the \texttt{chatml} template. 




\begin{table}[h!]
\centering
\begin{tabular}{ll}
\toprule
\textbf{Language Code} & \textbf{Language} \\
\midrule
ast & Asturian \\
ca & Catalan \\
de & German \\
el & Greek \\
es & Spanish \\
en & English \\
eu & Basque \\
fr & French \\
gl & Galician \\
it & Italian \\
nl & Dutch \\
pt & Portuguese \\
\bottomrule
\end{tabular}
\caption{Mapping from BCP-47 language codes to full language names.}
\label{tab:lang_code_map}
\end{table}

\newcommand{\FixWidthCell}[1]{\makebox[6em][c]{#1}}

\begin{table*}[p!]
  \centering
  \begin{tabular}{l r r r c}
    \toprule
       Pair & Count  & Source Len. ($\mu$ ± $\sigma$; med) & Target Len. ($\mu$ ± $\sigma$; med)& Cosine Sim. ($\mu$ ± $\sigma$; med) \\
    \midrule
    en-es  & 347479 & 1066±891; 896  & 1169±1007; 975 & 0.92±0.03; 0.93 \\
    en-fr  & 139350 & 899±420; 839   & 978±452; 915   & 0.90±0.03; 0.91 \\
    ca-en  &  63615 & 1318±1098; 1018 & 1230±1025; 958 & 0.92±0.03; 0.92 \\
    ca-es  &  56407 & 1136±973; 895  & 1162±992; 918  & 0.98±0.02; 0.98 \\
    es-fr  &  31301 & 915±448; 855   & 918±454; 856   & 0.95±0.02; 0.96 \\
    en-pt  &  24596 & 1005±432; 948  & 1037±449; 978  & 0.90±0.03; 0.90 \\
    es-pt  &  13952 & 1152±504; 1102 & 1096±484; 1043 & 0.97±0.02; 0.97 \\
    de-fr  &   7531 & 1017±508; 947  & 986±491; 922   & 0.92±0.03; 0.93 \\
    de-en  &   6587 & 1032±505; 965  & 927±460; 862   & 0.88±0.03; 0.88 \\
    es-eu  &   6272 & 780±725; 641   & 710±660; 581   & 0.92±0.03; 0.92 \\
    en-it  &   6128 & 910±496; 832   & 969±553; 884   & 0.92±0.03; 0.93 \\
    en-gl  &   5858 & 1314±700; 1236 & 1375±738; 1285 & 0.91±0.03; 0.91 \\
    fr-pt  &   5829 & 984±491; 907   & 922±458; 850   & 0.95±0.02; 0.96 \\
    es-gl  &   5618 & 1387±831; 1332 & 1319±795; 1271 & 0.98±0.01; 0.99 \\
    en-eu  &   4408 & 944±667; 774   & 938±678; 767   & 0.88±0.03; 0.88 \\
    fr-it  &   3863 & 951±428; 904   & 932±402; 886   & 0.95±0.02; 0.96 \\
    de-es  &   2339 & 992±467; 916   & 947±427; 872   & 0.93±0.03; 0.93 \\
    ca-fr  &   1806 & 915±489; 874   & 943±513; 903   & 0.96±0.02; 0.96 \\
    es-it  &   1513 & 1149±845; 966  & 1144±861; 957  & 0.96±0.02; 0.97 \\
    de-it  &    703 & 1124±419; 1082 & 1047±390; 1000 & 0.93±0.02; 0.94 \\
    eu-fr  &    621 & 499±387; 431   & 553±393; 486   & 0.89±0.04; 0.90 \\
    fr-nl  &    509 & 905±504; 892   & 909±503; 897   & 0.93±0.03; 0.94 \\
    el-fr  &    435 & 913±386; 912   & 900±383; 890   & 0.88±0.04; 0.89 \\
    el-en  &    433 & 916±399; 912   & 850±371; 843   & 0.87±0.03; 0.87 \\
    en-nl  &    417 & 1000±533; 977  & 1089±613; 1072 & 0.90±0.03; 0.91 \\
    it-pt  &    369 & 952±282; 954   & 891±271; 881   & 0.96±0.02; 0.96 \\
    de-pt  &    316 & 1040±342; 1038 & 892±311; 887   & 0.94±0.02; 0.94 \\
    ca-pt  &    308 & 1059±589; 910  & 1027±562; 898  & 0.96±0.03; 0.96 \\
    ca-it  &    273 & 1000±851; 822  & 1014±881; 799  & 0.96±0.02; 0.97 \\
    ca-eu  &    112 & 884±696; 743   & 848±627; 708   & 0.93±0.02; 0.93 \\
    ca-de  &     59 & 1122±953; 904  & 1228±1139; 976& 0.93±0.03; 0.94 \\
    gl-pt  &     47 & 1811±1195; 1446& 1755±1161; 1418& 0.98±0.02; 0.98 \\
    ast-en &     41 & 636±510; 573   & 633±527; 502   & 0.88±0.04; 0.89 \\
    ca-gl  &     33 & 1307±972; 1032 & 1282±911; 995  & 0.96±0.03; 0.96 \\
    es-nl  &     13 & 2887±1926; 2651& 3045±2282; 1747& 0.93±0.02; 0.94 \\
    fr-gl  &     13 & 2759±3853; 1156& 2652±3638; 1106& 0.94±0.04; 0.94 \\
    de-eu  &     10 & 1048±678; 873  & 951±559; 905   & 0.89±0.02; 0.89 \\
    it-nl  &      9 & 1048±894; 1130 & 1074±896; 1157 & 0.91±0.05; 0.92 \\
    el-es  &      8 & 2081±1300; 1817& 2128±1417; 1723& 0.88±0.04; 0.89 \\
    de-nl  &      6 & 1497±939; 1538 & 1442±834; 1553 & 0.91±0.04; 0.92 \\
    gl-it  &      5 & 870±342; 878   & 958±452; 898   & 0.95±0.02; 0.96 \\
    eu-pt  &      4 & 663±235; 629   & 631±238; 563   & 0.92±0.04; 0.94 \\
    ast-ca &      4 & 67±15; 69      & 65±15; 63      & 0.98±0.01; 0.99 \\
    ca-el  &      3 & 1959±1181; 1287& 2020±1111; 1470& 0.87±0.05; 0.86 \\
    ca-nl  &      3 & 2997±1075; 3260& 3293±1452; 3238& 0.93±0.03; 0.93 \\
    de-gl  &      2 & 2668±2598; 2668& 2498±2487; 2497& 0.91±0.00; 0.91 \\
    eu-it  &      2 & 473±172; 472   & 484±74; 484   & 0.86±0.01; 0.86 \\
    eu-gl  &      1 & 655±–; 655     & 662±–; 662     & 0.95±–; 0.95 \\
    \bottomrule
  \end{tabular}
  \caption{Main statistics of the \acadtrain\ set, shown only for one direction.}
  \label{tab:summary_train}
\end{table*}


\begin{table*}[!ht]
\centering
\begin{tabular}{l r r r c}
\toprule
\textbf{Pair} & \textbf{Count} & \textbf{Src Len. ($\mu$ ± $\sigma$; med)} & \textbf{Tgt Len. ($\mu$ ± $\sigma$; med)} & \textbf{Cosine Sim. ($\mu$ ± $\sigma$; med)} \\
\midrule

en–es  & 2161 & 1102±809; 909   & 1203±879; 993  & 0.93±0.03; 0.93 \\
en–fr  & 333  & 897±392; 836    & 969±429; 902   & 0.91±0.03; 0.92 \\
ca–en  & 210  & 1190±949; 966   & 1115±904; 876  & 0.92±0.03; 0.92 \\
ca–es  & 188  & 1290±1123; 915  & 1317±1132; 945 & 0.98±0.02; 0.98 \\
es–fr  & 46   & 798±247; 775    & 795±244; 780   & 0.95±0.03; 0.95 \\
en–pt  & 34   & 1037±408; 970   & 1065±416; 1036 & 0.90±0.03; 0.90 \\

\midrule
\textbf{Overall} & \textbf{2972} & \textbf{1091±779; 904} & \textbf{1169±832; 969} & \textbf{0.93±0.03; 0.93} \\
\bottomrule
\end{tabular}
\caption{Main statistics of the \acadbench\ set, shown only for one direction.}
\label{tab:summary_test}
\end{table*}

\definecolor{customimstart}{HTML}{f9f4e6} 
\definecolor{customimend}{HTML}{f2fbf6} 

\definecolor{colorhex1}{HTML}{f9f4e6} 
\definecolor{colorhex2}{HTML}{f2fbf6} 
\definecolor{colorhex3}{HTML}{e6f1f9}

\begin{table*}[!ht]
\centering
\footnotesize
\begin{tabularx}{\textwidth}{lX}
\toprule
\multirow{11}{*}{\textbf{User}} 
  & \highlight{customimstart}{\textbf{\texttt{<|im\_start|>user}}} \\
  & \textbf{Translate the following text from English to Spanish.} \\
  & \textbf{English:} With the purpose of analyzing women’s perceptions and classifying their modes of understanding a positive human papillomavirus (HPV+) test, we conducted 38 in‑depth interviews with women who had received an HPV diagnosis (normal and abnormal Pap smear), screened in Jujuy’s public health system in 2016. A typology based on women’s understandings of the result was developed: 1) understanding; 2) lack of understanding; a) underestimation; b) overestimation; c) confusion. The interviewees who experienced confusion over the results reported contradictory perceptions in relation to a positive HPV test and its severity; those who underestimated it tended to mention the absence of symptoms and expressed little concern over the result; while those who overestimated it considered themselves sick and described concern, narrating a biographical disruption and physical pain. These findings confirm the need to improve the delivery of results and the provision of information in order to decrease psychosocial impact and increase follow‑up adherence in HPV‑positive women. 
  \\
  & \textbf{Spanish:}  \highlight{customimend}{\textbf{\texttt {<|im\_end|>}}} \\
\midrule
\multirow{11}{*}{\textbf{Model}} 
  & \highlight{customimstart}{\textbf{\texttt{<|im\_start|>assistant}}} \\
  & Con el propósito de analizar las percepciones de las mujeres y clasificar sus modos de comprensión de un resultado positivo de virus del papiloma humano (VPH+), en 2016 realizamos 38 entrevistas en profundidad a mujeres con diagnóstico de VPH (citología normal y anormal) detectado en el sistema público de salud de Jujuy. Se elaboró una tipología basada en la comprensión del resultado por parte de las mujeres: 1) comprensión; 2) falta de comprensión; a) subestimación; b) sobreestimación; c) confusión. Las entrevistadas que experimentaron confusión informaron percepciones contradictorias sobre el VPH+ y su gravedad; quienes lo subestimaron tendían a mencionar la ausencia de síntomas y mostraron poca preocupación; mientras que aquellas que lo sobreestimaron se consideraban enfermas, describían preocupación, narrando una ruptura biográfica y dolor físico. Estos hallazgos confirman la necesidad de mejorar la entrega de resultados y la provisión de información para disminuir el impacto psicosocial y aumentar la adherencia al seguimiento en mujeres con VPH positivo.\highlight{customimend}{\textbf{\texttt{<|im\_end|>}}} 
  \\
\bottomrule
\end{tabularx}
\caption{Example of an instance from \acadtrain\ dataset formatted using the ChatML template.
}
\label{tab:instruction_template}
\end{table*}

\section{\acadbench\ Domain Distribution}
\label{app:test_dist}

Table~\ref{tab:domain_classifier_description} provides descriptions of the domains predicted by NVIDIA's Multilingual Domain Classifier. 

\begin{figure}[!ht]
\centering
  \includegraphics[width=\linewidth]{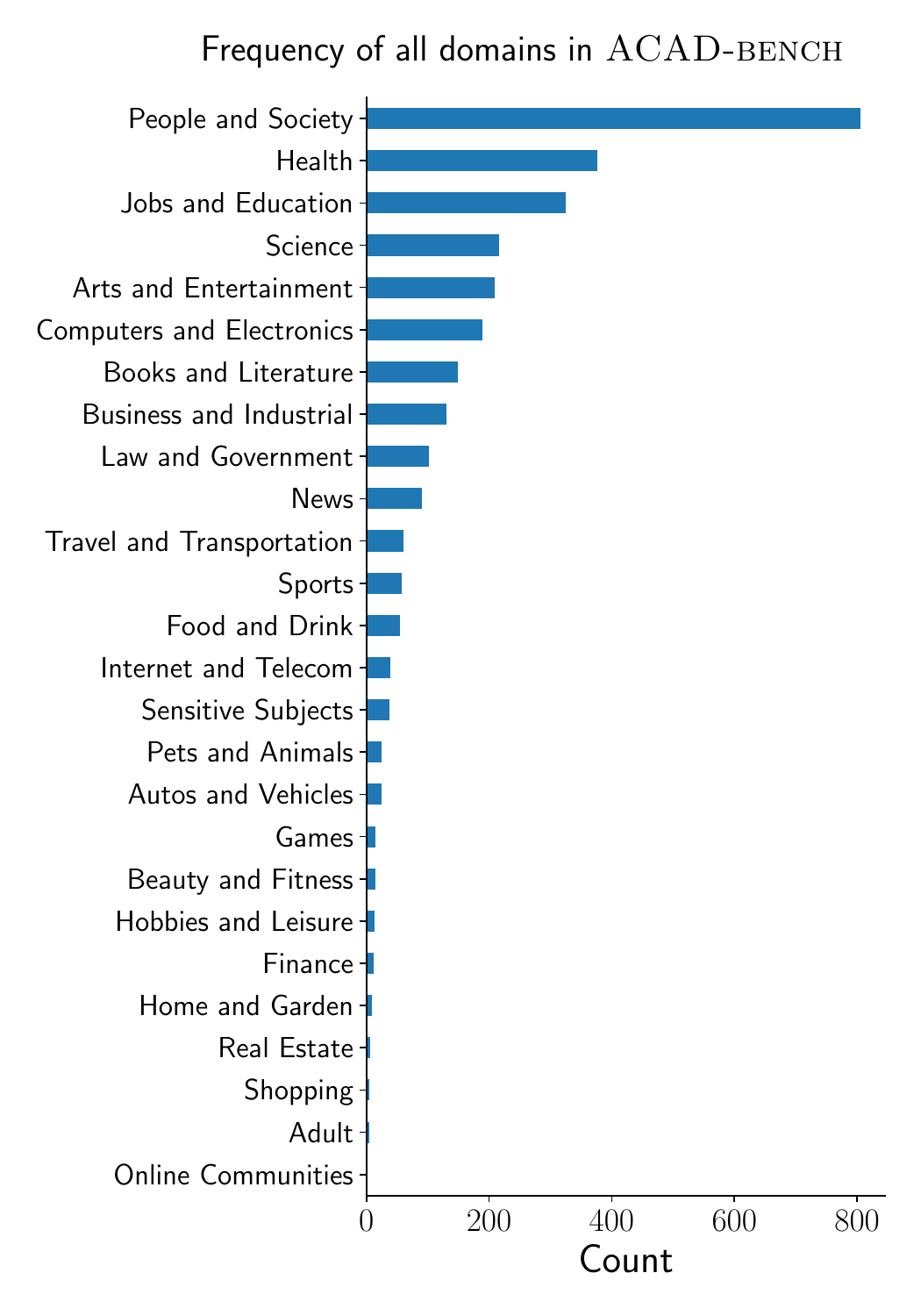}
  \caption{Bar plot showing all the domains in \acadbench.}\label{fig:freq_all_domains}
\end{figure}

Figure \ref{fig:freq_all_domains} illustrates that \acadbench's content is concentrated in a handful of major areas. The four largest domains: People and Society (806 instances, $\approx$21\%), Health (376, $\approx$9.8\%), Jobs and Education (325, $\approx$8.5\%), and Science (216, $\approx$5.7\%), together account for just over 45\% of all examples. A second group of domains: Arts and Entertainment (209), Computers and Electronics (189), Books and Literature (149), and Business and Industrial (130), each represent between 3 and 5\% of the benchmark.

The remaining 17 domains span a variety of topics: from Law and Government (102) and News (90) down to the smallest categories such as Real Estate (6), Shopping (5), Adult (4), and Online Communities (2), and collectively make up the final $\approx$30\% of instances. This breadth ensures that \acadbench\ covers both widely studied areas and more specialized subjects.

\begin{table*}[ht]
\centering
\begin{tabular}{p{4.2cm}p{10.5cm}}
\toprule
\textbf{Domain Class} & \textbf{Description} \\
\midrule
Adult & Sexual content, pornography, or age-restricted material \\
Arts\_and\_Entertainment & Music, movies, theater, celebrities, pop culture \\
Autos\_and\_Vehicles & Cars, motorbikes, vehicle news and reviews \\
Beauty\_and\_Fitness & Skincare, cosmetics, wellness, workout routines \\
Books\_and\_Literature & Novels, literary criticism, poetry, book reviews \\
Business\_and\_Industrial & Enterprise, corporate, manufacturing, B2B topics \\
Computers\_and\_Electronics & Hardware, software, tech news, consumer gadgets \\
Finance & Banking, investing, personal finance, stock markets \\
Food\_and\_Drink & Recipes, restaurants, food culture, drinks \\
Games & Video games, board games, eSports, gaming culture \\
Health & Medical topics, mental health, wellness, diseases \\
Hobbies\_and\_Leisure & DIY, crafts, hobbies, leisure activities \\
Home\_and\_Garden & Home improvement, gardening, decor \\
Internet\_and\_Telecom & ISPs, web platforms, telecommunications \\
Jobs\_and\_Education & Career guidance, job listings, academic topics \\
Law\_and\_Government & Legislation, public policy, political topics \\
News & Journalism, current events, news reporting \\
Online\_Communities & Forums, social platforms, user communities \\
People\_and\_Society & Culture, social issues, demographics \\
Pets\_and\_Animals & Pet care, wildlife, zoology topics \\
Real\_Estate & Property listings, housing market, realty advice \\
Science & Research, scientific articles, STEM topics \\
Sensitive\_Subjects & Controversial or delicate content (e.g. abuse, violence) \\
Shopping & E-commerce, product reviews, retail \\
Sports & Athletic events, scores, sports commentary \\
Travel\_and\_Transportation & Tourism, transit, travel guides \\
\bottomrule
\end{tabular}
\caption{Domain class descriptions for NVIDIA’s multilingual domain classifier, based on manual inspection of sample instances.}
\label{tab:domain_classifier_description}
\end{table*}

\section{Inference via API}
\label{app:api_inference}

This appendix section describes our end-to-end inference pipeline across different model families. We outline the prompt templates used, highlight challenges in obtaining strictly formatted outputs, and specify the interfaces and settings leveraged for each model, whether via remote APIs or local deployments.

\subsection{Challenges with Model Output Formatting}
When using the prompt template defined in Table~\ref{tab:instruction_template}, we observe that certain conversational models frequently prepend or append unsolicited comments instead of returning a raw translation. Examples of such extraneous text include:
\begin{itemize}
  \item “Here's the translation from \{src\} to \{tgt\}:”
  \item “Breakdown of choices: …”
  \item “Let me know if you'd like any specific parts clarified or alternative phrasing considered!”
\end{itemize}


\begin{figure*}[h]
  \centering
\begin{tcolorbox}[
  colback=gray!5,
  colframe=black,
  title=Prompt template used for inference,
  fonttitle=\bfseries,
  fontupper=\small\ttfamily,
  arc=4mm,
  boxrule=1pt,
  width=0.85\textwidth,
]
\begin{Verbatim}[commandchars=\\\{\}]
Only output the translation, only one option, no explanations or anything else:
Translate the following text from \textbf{\{src. language\}} to \textbf{\{tgt. language\}}. \\
\textbf{\{src. language\}}: \textbf{\{src. text\}} \\
\textbf{\{tgt. language\}}:
\end{Verbatim}
\end{tcolorbox}

  \caption{Adapted prompt for inference with models that cannot follow the original template format. Implemented to force a single, clean translation.}
  \label{fig:adapted_template}
\end{figure*}

We attribute this behavior to the conversational datasets used to train these models, which encourage them to add explanations, multiple options, or task descriptions, and make them unable to fully understand the specified format in the prompt. To enforce a strict “translation-only” output, we design an \emph{adapted} prompt (Figure~\ref{fig:adapted_template}) that explicitly instructs the model to produce only the translated text, with no additional prefixes, suffixes, or comments. Models that reliably adhere to the original instruction template are evaluated using the unmodified prompt from Table~\ref{tab:instruction_template}.

\subsection{API and Deployment Details}
For each model family, we use the following inference interfaces and settings:

\paragraph{OpenAI API}
We access closed-source OpenAI models via the official OpenAI API\footnote{\url{https://openai.com/api/}} for the following models; \gptmini\footnote{Retrieved July 21, 2025, via OpenAI API from \href{https://platform.openai.com/docs/models/gpt-4.1-mini}{https://platform.openai.com/docs/models/gpt-4.1-mini}}, \gptnano\footnote{Retrieved July 21, 2025, via OpenAI API from \href{https://platform.openai.com/docs/models/gpt-4.1-nano}{https://platform.openai.com/docs/models/gpt-4.1-nano}}. Since these endpoints do not accept custom generation hyperparameters, we rely on the default settings. Notably, both \gptmini{} and \gptnano{} follow the original “translation-only” instruction without requiring the adapted template.

\paragraph{Google API}
Inference on Google's \geminitwofive{}\footnote{Retrieved July 24, 2025, via Google AI API from \href{https://cloud.google.com/vertex-ai/generative-ai/docs/models/gemini/2-5-flash}{https://cloud.google.com/vertex-ai/generative-ai/docs/models/gemini/2-5-flash}}, \geminitwo{}\footnote{Retrieved July 24, 2025, via Google AI API from \href{https://cloud.google.com/vertex-ai/generative-ai/docs/models/gemini/2-0-flash}{https://cloud.google.com/vertex-ai/generative-ai/docs/models/gemini/2-0-flash}}, and \gemmathree{} is performed through the Google AI API\footnote{\url{https://ai.google.dev/}}. These models likewise do not expose generation hyperparameters. All three models comply with the original prompt and do not require any template adaptations.

\paragraph{Local Hugging Face Deployments}
For open-source models, including \salamandra\ (2B, 7B, and their fine-tuned variants) and \llamathree, we load model checkpoints in Hugging Face format and run inference locally. In these cases we use beam search decoding, setting the beam size to 5.


\section{Win Margin}\label{app:win_rate}

To compare two systems, we define a \emph{win} when the difference in \bleu\ scores between the alternative system and the baseline system exceeds a threshold of 5 points. Formally, let $s_i^{(A)}$ and $s_i^{(B)}$ denote the \bleu\ scores for source text $i$ from the alternative system $A$ and the baseline system $B$, respectively. The \textit{delta} is computed as:
\[
\Delta_i = s_i^{(A)} - s_i^{(B)}.
\]
We define a win as $\Delta_i > 1$, and a loss as $\Delta_i < -1$. Ties or negligible differences ($|\Delta_i| \leq 5$) are ignored. The \textbf{win rate margin} is then computed as:
\[
\text{WinMargin} = 100 \cdot \frac{N_{\text{wins}} - N_{\text{losses}}}{N_{\text{valid}}}
\]
where $N_{\text{wins}}$ is the number of instances with $\Delta_i > 5$, $N_{\text{losses}}$ is the number of instances with $\Delta_i < - 5$, and $N_{\text{valid}} = N_{\text{wins}} + N_{\text{losses}}$ is the number of comparisons where a win or loss occurred.

To analyze performance with respect to length, we group the source text into $Q$ quantiles based on their token length and report the win margin per quantile.


\begin{figure*}[t]
  \centering
  \begin{tcolorbox}[
      enhanced,
      colback=gray!2,        
      colframe=gray!100,      
      toprule=0.8pt, leftrule=0.8pt, rightrule=0.8pt, bottomrule=0.8pt,
      title={Spanish \(\,\|\)\ English}, 
      coltitle=white,        
      colbacktitle=gray!80,  
      fonttitle=\bfseries,
      sharp corners,
  ]

    Las denominadas estrategias de gamificación favorecen el aprendizaje significativo al incrementar la motivación del alumnado. En la materia inmunología, del grado en farmacia se abordan contenidos nuevos de gran complejidad por ello, la inclusión de actividades lúdicas podría mejorar los resultados académicos. En el presente trabajo se diseñó una actividad cooperativa en la que los estudiantes debían planificar, diseñar y ejecutar un juego que permitiera a estudiantes de grado repasar inmunología. Para evaluar la utilidad de los juegos creados, se diseñó una sesión de juego en la que participaron los alumnos y cinco graduados en Farmacia y tres docentes del grado. Como observadores externos. Los resultados obtenidos fueron satisfactorios siendo la calificación promedio del grupo 8,1. Además, para el 76,7\% de los participantes, crear el juego permitió comprender mejor la materia. Por ello, el diseño de juegos de repaso constituye una buena herramienta para mejorar el abordaje de la asignatura.

    \tcblower
    
    Gamification strategies improve meaningful learning by increasing student motivation. In the subject of Immunology, of the Pharmacy Degree, new content of great complexity is addressed, therefore, the inclusion of games could improve academic results. In the present work, a cooperative activity was designed in which students had to plan, design and execute a game that allowed undergraduate students to review immunology. To evaluate the usefulness of the created games, a game session was designed in which the students and five pharmacy graduates and three degree teachers participated. as outside observers. The results obtained were satisfactory, with the average rating of the group being 8.1. In addition, for 76.7\% of the participants, creating the i-game allowed them to better understand the subject. For this reason, the design of revision games constitutes a good tool to improve the approach to the subject.

    \end{tcolorbox}
  \caption{\acadtrain{} translation pair in Spanish and English.}
  \label{fig:example1}
\end{figure*}

\begin{figure*}[t]
  \centering
  \begin{tcolorbox}[
      enhanced,
      colback=gray!2,        
      colframe=gray!100,      
      toprule=0.8pt, leftrule=0.8pt, rightrule=0.8pt, bottomrule=0.8pt,
      title={English \(\,\|\)\ Basque}, 
      coltitle=white,        
      colbacktitle=gray!80,  
      fonttitle=\bfseries,
      sharp corners,
  ]
In this paper we analyze, first, the general characteristics of the Spanish development model of the last thirty years as part of its role in the process of global accumulation of capital. Secondly we discover, from the analysis of the changes in production in a paradigmatic sector (the software industry). We describe how the production model has been shaped through organizational and commercial corporate strategies. Thirdly we analyze the possibilities for a future development in the new global economic framework, choosing a perspective different from corporate rentability strategies.
    \tcblower
    
    Artikulu honetan aztertzen dira, batetik, azken hogeita hamar urteetako espainiar garapenaren ezaugarri orokorrak handitutako kapitalaren metaketa prozesuan era globalean bete duen papera aintzat hartuz. Bestetik, honako hau deskubritu dugu: arlo paradigmatiko batean (software-aren industrian) jasandako produkzio aldaketen analisitik abiatuz, produkzio eredu hori nola konfiguratu den antolaketaeta merkatu-estrategien bidez. Hirugarrenik, ekonomiaren marko berrian etorkizunari begira garapen aukerak aztertzen ditugu, enpresa-errentagarritasun estrategien beste norabide batetik abiatuz

    \end{tcolorbox}
  \caption{\acadtrain{} translation pair in English and Basque.}
  \label{fig:example4}
\end{figure*}

\begin{figure*}[t]
  \centering
  \begin{tcolorbox}[
      enhanced,
      colback=gray!2,        
      colframe=gray!100,      
      toprule=0.8pt, leftrule=0.8pt, rightrule=0.8pt, bottomrule=0.8pt,
      title={French \(\,\|\)\ English}, 
      coltitle=white,        
      colbacktitle=gray!80,  
      fonttitle=\bfseries,
      sharp corners,
  ]
La figure juridique du protecteur prend ses racines dans l'Empire romain, a été reprise par le christianisme, associée au concept de charité et récupérée dans l'Europe moderne dans son triple aspect de gardien, médiateur et défenseur. Dans l'administration espagnole, la figure du protecteur des Indiens est la plus connue, grâce aux nombreuses études sur le statut juridique de l'Indien. Le `Protecteur des étrangers', en revanche, est pratiquement inconnu. Cette contribution examine son évolution et la pratique de la protection. Il affirme que le `Protecteur des étrangers' a pu aider à la gestion d'une altérité qui est passée en peu de temps d'un espace péninsulaire réduit à une dimension mondiale. La pratique de la protection, cependant, a été canalisée par une liste technique de subventions si complexe, et un ensemble de mécanismes sociaux non institutionnalisés si large, qu'il y avait en fait une multiplicité de figures agissant comme protecteurs.

    \tcblower
    
    The legal figure of the protector has its roots in the Roman Empire; it was taken up by the Christianity associated with the concept of charity and recovered in modern Europe in its triple aspect of guardian, mediator and defender. In the Spanish administration, the figure of the protector of the indians is the best known, thanks to the numerous studies on the legal status of the indigenous. The `Protector of Foreigners', on the other hand, is virtually unknown. This contribution examines its evolution and the practice of protection. It argues that the `Protector of Foreigners' could have contributed to the management of an otherness that moved, in a short time frame, from a reduced peninsular space to a global dimension. The practice of protection, however, was channelled by such a complex, technical list of subsidies, and such a broad set of non-institutionalised social mechanisms, that there was, in fact, an abundance of figures acting as protectors.

    \end{tcolorbox}
  \caption{\acadtrain{} translation pair in French and English.}
  \label{fig:example3}
  \vspace{-12pt}
\end{figure*}

\begin{figure*}[t]
  \centering
  \begin{tcolorbox}[
      enhanced,
      colback=gray!2,        
      colframe=gray!100,      
      toprule=0.8pt, leftrule=0.8pt, rightrule=0.8pt, bottomrule=0.8pt,
      title={Catalan \(\,\|\)\ German}, 
      coltitle=white,        
      colbacktitle=gray!80,  
      fonttitle=\bfseries,
      sharp corners,
  ]

La Guerra de Successió generà tota una publicística textual, divulgada a través d'impresos i manuscrits amb intervenció de l'oralitat, destinada a conformar una opinió pública d'acord amb la ideologia i els propòsits dels poders que aspiraven a ser dominants, a enaltir el fervor popular i a desacreditar els afectes al bàndol contrari. En aquestes notes, s'intenta esbossar el fenomen des de les línies austriacistes.

    \tcblower
    
    Der Spanische Erbfolgekrieg schuf eine umfangreiche Textpublizistik, die über Drucklegungen und öffentlich verlesene Manuskripte verbreitet wurde und dazu dienen sollte, die öffentliche Meinung gemäß Ideologie und Absichten der um die Vorherrschaft kämpfenden Mächte zu gestalten, die Begeisterung des Volkes zu schüren und die Anhänger der Gegenseite zu diskreditieren. In diesen Notizen wird versucht, das Phänomen von der Warte der Habsburger aus zu skizzieren.

    \end{tcolorbox}
  \caption{\acadtrain{} translation pair in Catalan and German.}
  \label{fig:example2}
\end{figure*}

\begin{figure*}[t]
  \centering
  \begin{tcolorbox}[
      enhanced,
      colback=gray!2,        
      colframe=gray!100,      
      toprule=0.8pt, leftrule=0.8pt, rightrule=0.8pt, bottomrule=0.8pt,
      title={Portuguese \(\,\|\)\ Spanish}, 
      coltitle=white,        
      colbacktitle=gray!80,  
      fonttitle=\bfseries,
      sharp corners,
  ]
Introdução. O reconhecimento da percepção do usuário sobre a qualidade do serviço de emergência por meio do modelo avaliativo Service Performance permite a orientação de ações para melhorar os serviços prestados. Objetivo. Determinar a percepção da qualidade do serviço de emergência de uma Empresa Social do Estado em Boyacá. Método. Foi realizada uma pesquisa descritiva transversal, com uma amostra composta por 108 pessoas que receberam cuidados no serviço de emergência durante o segundo semestre de 2017. Para a coleta da informação foi usado Service Performance, que tem 22 questões organizadas em cinco dimensões que permitem avaliar a qualidade do serviço de acordo com a percepção do usuário. Resultados. A percepção dos usuários sobre a qualidade do serviço de emergência, em cada uma das cinco dimensões, obteve uma pontuação média geral de 4,09, sendo a dimensão de segurança a maior, com uma média de 4,23, seguida da empatia, com 4,18, a capacidade da resposta, com 4,14, a confiabilidade, com 4,09, e a menor foram os elementos tangíveis, com 3,79. Conclusões. Conhecer a percepção da qualidade do serviço de emergência permite desenvolver ações, estabelecer confiança e monitorar a qualidade do serviço.

    \tcblower
    
    Introducción. El reconocimiento de la percepción del usuario de la calidad del servicio de urgencias mediante el modelo evaluativo Service Performance, permite la orientación de acciones para mejorar los servicios prestados. Objetivo. Determinar la percepción de la calidad de la atención del servicio de urgencias de una Empresa Social del Estado en Boyacá. Método. Se hizo una investigación descriptiva de diseño transversal, con una muestra conformada por 108 usuarios que recibieron atención en el servicio de urgencias durante el segundo semestre del año 2017. Para la recolección de la información, se utilizó la encuesta Service Performance, la cual consta de 22 preguntas organizadas en cinco dimensiones que permiten evaluar la calidad del servicio según la percepción del usuario. Resultados. La percepción de los usuarios sobre la calidad del servicio de urgencias, en cada una de las cinco dimensiones, obtuvo un puntaje promedio general de 4,09, siendo la dimensión de seguridad la más alta, con un promedio de 4,23, seguida de la empatía, con 4,18, la de capacidad de respuesta, con 4,14, la de fiabilidad, con 4,09 y, la más baja, los elementos tangibles, con 3,79. Conclusiones. El conocer la percepción de la calidad del servicio de urgencias permite desarrollar acciones, establecer confianza y hacer el seguimiento de la calidad del servicio.

    \end{tcolorbox}
  \caption{\acadtrain{} translation pair in Portuguese and Spanish.}
  \label{fig:example5}
  \vspace{-12pt}
\end{figure*}

\begin{figure*}[t]
  \centering
  \begin{tcolorbox}[
      enhanced,
      colback=gray!2,        
      colframe=gray!100,      
      toprule=0.8pt, leftrule=0.8pt, rightrule=0.8pt, bottomrule=0.8pt,
      title={English \(\,\|\)\ Spanish}, 
      coltitle=white,        
      colbacktitle=gray!80,  
      fonttitle=\bfseries,
      sharp corners,
  ]
This paper presents a compilation of the characteristics of Crocus sativus L., saffron, a species prized since antiquity. The paper considers botanical and biochemical as well as data on the cultivation and marketing basics.

    \tcblower
    
    El presente trabajo muestra una recopilación sobre las características de Crocus sativus L., el azafrán, una especie, muy apreciada desde la antigüedad. Se consideran aspectos básicos botánicos y bioquímicos así como datos sobre el cultivo, y su comercialización.

    \end{tcolorbox}
  \caption{\acadtrain{} translation pair in English and Spanish.}
  \label{fig:example6}
\end{figure*}

\begin{figure*}[t]
  \centering
  \begin{tcolorbox}[
      enhanced,
      colback=gray!2,        
      colframe=gray!100,      
      toprule=0.8pt, leftrule=0.8pt, rightrule=0.8pt, bottomrule=0.8pt,
      title={Catalan \(\,\|\)\ Italian}, 
      coltitle=white,        
      colbacktitle=gray!80,  
      fonttitle=\bfseries,
      sharp corners,
  ]
¿Què fa que les persones siguin eficaces al treball? En les darreres dècades, comparacions entre l'assoliment superior i promig mostren que les persones amb major acompliment al treball són aquelles que manifesten competències comportamentals, les quals estan relacionades amb el comportament emocional i social. Amb la integració de teories de diferents camps, aquesta tesi contribueix a l'actual debat sobre competències comportamentals facilitant una millor comprensió dels seus antecedents, resultats i medicions. La demanda persistent el mercat de treball de professionals amb habilitats comportamentals més altes subratlla la necessitat d'aprofundir en la investigació i esbrinar, a part de la formació protocolar, què més pot ajudar al desenvolupament d'aquestes competències. A partir de la teoria d'aprenentatge experimental, el primer estudi, basat en una mostra d'estudiants licenciats, revela que una sèrie d'activitats extracurriculars pròpies de la vida real semblen predir diferents clústers de competències comportamentals. El segon estudi avalua l'impacte que les competències comportamentals tenen sobre la forma en que els estudiants licenciats encaren la seva carrera professional i com això afecta la seva ocupabilitat. El tercer estudi té com objectiu medir amb eficàcia competències comportamentals amb finalitats avaluatives i de desenvolupament personal. En aquest estudi es desenvolupa un instrument de mesura amb la intenció de posar al dia, ampliar i superar algunes limitacions pròpies de les escales existents. Basat en una revisió de la literatura i en investigacions empíriques addicionals, es proposa un nou i exhaustiu marc competencial, i es desenvolupen i validen les corresponents escales de mesura. Aquesta investigació ofereix un novedós model de mesura per competències comportamentals i mostra la importància del seu desenvolupament ja a una edat primerenca de la vida per tal que ajudi als estudiants a aventurar-se amb èxit al mercat laboral

    \tcblower
    
    Cosa rende le persone efficaci nel loro lavoro? Negli ultimi anni una comparazione sistematica tra i performer migliori e quelli nella media ha evidenziato che le persone con le migliori prestazioni sono quelle che dimostrano competenze trasversali, ovvero competenze legate a comportamenti di comprensione e gestione di emozioni e relazioni. Attraverso l'integrazione di teorie provenienti da diversi ambiti scientifici, questa tesi contribuisce all'attuale dibattito sulle competenze trasversali attraverso l'approfondimento di fattori che facilitano lo sviluppo di queste competenze, una miglior comprensione dei loro effetti e delle loro modalità di misurazione. La persistente discrepanza in termini di competenze trasversali tra domanda e offerta nel mondo del lavoro sottolinea la necessità di indagare maggiormente i fattori che possono contribuire allo sviluppo di queste competenze. Basandosi sulla teoria dell'apprendimento esperienziale, il primo studio condotto su un campione di studenti universitari, fa luce sull'impatto di una serie di attività extracurriculari su diverse tipologie di competenze trasversali. Il secondo studio valuta l'influenza delle competenze trasversali sul modo in cui i neolaureati orientano la propria carriera e le relative conseguenze in termini di employability. I risultati ottenuti indicano che le competenze trasversali risultano critiche non solo per il successo nella carriera di professionisti, come precedentemente enfatizzato dalla letteratura, ma anche per un efficace orientamento alla carriera degli studenti e una conseguente maggiore occupabilità. Il terzo studio propone un efficace strumento di misurazione delle competenze trasversali, volto ad aggiornare, ampliare e a superare i limiti dei modelli esistenti. Basandosi sull'analisi della letteratura e su un'addizionale ricerca empirica, lo studio propone un nuovo modello di competenze trasversali e presenta lo sviluppo e la validazione delle relative scale. Questa ricerca fornisce un modello innovativo di misurazione delle competenze trasversali e mostra l'importanza del loro sviluppo sin dalle prime fasi per guidare con successo gli studenti verso la loro avventura nel mondo del lavoro.

    \end{tcolorbox}
  \caption{\acadtrain{} translation pair in Catalan and Italian.}
  \label{fig:example7}
\end{figure*}

\begin{algorithm*}[!ht]
\begin{algorithmic}[1]
  \Procedure{SampleTestSet}{$\mathit{train\_data},\,\mathit{candidates}$}
    \ForAll{lang‐pair $p$ in $\mathit{train\_data}$}
      \State $\mathit{count}[p] \gets \bigl|\mathit{train\_data}[p]\bigr|$
    \EndFor

    \State $P \gets \{\,p \mid \mathit{count}[p] \ge 1000\}$

    \State $\mathit{cand} \gets [\,x \in \mathit{candidates}\mid \text{langPair}(x)\in P\,]$

    \For{$i = 1$ \textbf{to} $|\mathit{cand}|$}
      \State $\mathit{weights}[i] \gets \mathit{count}\bigl[\text{langPair}(\mathit{cand}[i])\bigr]$
    \EndFor

    \State $N \gets \min\bigl(3000,\,|\mathit{cand}|\bigr)$
    \State $\mathit{probs}\gets \text{Normalize}(\mathit{weights})$
    \State $\mathit{idxs}\gets \text{SampleWithoutReplacement}(1\,\dots,|\mathit{cand}|;N,\mathit{probs})$
    \State $\mathit{sampled}\gets [\,\mathit{cand}[i]\mid i\in\mathit{idxs}\,]$

    \State $\mathit{groups}\gets \text{GroupBy}(\mathit{sampled},\text{langPair})$
    \State $\mathit{valid\_pairs}\gets \{\,p \mid |\mathit{groups}[p]|\ge10\}$

    \State $\mathit{test\_instances}\gets [\,]$
    \ForAll{lang‐pair $p=(s,t)$ in $\mathit{valid\_pairs}$}
      \ForAll{example $x$ in $\mathit{groups}[p]$}
        \State append $(s\to t; x)$ to $\mathit{test\_instances}$
        \State append $(t\to s; x)$ to $\mathit{test\_instances}$
      \EndFor
    \EndFor

    \State \Return $\mathit{test\_instances}$
  \EndProcedure
\end{algorithmic}
\caption{\acadbench\ Sampling Pseudocode}
\label{alg:testset_sampling}
\end{algorithm*}

\section{Results}\label{app:results}

\subsection{Results test set}

In Tables~\ref{tab:results_xx_en}, \ref{tab:results_en_xx}, \ref{tab:results_xx_es}, and~\ref{tab:results_es_xx}, we show detailed results per language pair for  xx→en, en→xx, xx→es and es→xx directions, respectively, evaluated on \acadbench.

\begin{table*}[!ht]
\centering
\small
\tabcolsep 9pt
\scalebox{0.97}{%
\begin{tabular}{llrrrrr}
\toprule
Pair & Model & {\small d-}\textsc{BLEU} & \textsc{BP} & \textsc{Blonde} & \textsc{Comet} & \textsc{Comet-}\textsc{\scriptsize Kiwi} \\
\midrule
{es→en} 
        & {\colorbox{colorhex1}{\makebox[4.2cm][l]{\gptmini\ }}}  & 48.06	& 1.00  & 0.60 & 0.84 & 0.78  \\
        & {\colorbox{colorhex1}{\makebox[4.2cm][l]{\gptnano\ }}}  & 42.81 	& 0.97 & 0.56 & 0.84 & 0.78  \\
        & {\colorbox{colorhex1}{\makebox[4.2cm][l]{\geminitwo\ }}}	 & 51.27&1.0&0.62&0.84&0.77 \\
        & {\colorbox{colorhex1}{\makebox[4.2cm][l]{\geminitwofive\ }}} & 46.99&0.98&0.6&0.84&0.77 \\
        & {\colorbox{colorhex2}{\makebox[4.2cm][l]{\llamathree$^\dagger$ }}} & 45.12&0.99&0.58&0.84&0.77 \\
        & {\colorbox{colorhex2}{\makebox[4.2cm][l]{\gemmathree$^\dagger$ }}} & 48.83&0.99&0.61&0.84&0.77 \\
        & {\colorbox{colorhex3}{\makebox[4.2cm][l]{\madlad$^\dagger$ }}} & 35.83&0.8&0.48&0.81&0.78 \\
        \cdashlinelr{2-7}
        & {\small \salamandratwo$^\dagger$} & 37.94	& 0.92 & 0.52 & 0.82 & 0.76 \\
        & \quad + {\small \acadtrain\ } & 51.28 & 1.00 & 0.62 & 0.84  & 0.76  \\
        \cdashlinelr{2-7}
        & {\small \salamandraseven$^\dagger$ } & 47.72 & 0.99 & 0.59 & 0.84  & 0.76  \\
        & \quad + {\small \acadtrain\ } & 53.06  & 1.00 & 0.63   & 0.84  & 0.76 \\
\cdashlinelr{1-7}
{pt→en}  & {\colorbox{colorhex1}{\makebox[4.2cm][l]{\gptmini\ }}} &  48.44&0.99&0.62&0.84&0.75 \\
        & {\colorbox{colorhex1}{\makebox[4.2cm][l]{\gptnano\ }}} &  44.10&0.97&0.56&0.84&0.76 \\
        & {\colorbox{colorhex1}{\makebox[4.2cm][l]{\geminitwo\ }}} & 51.09&1.0&0.63&0.83&0.75  \\
        & {\colorbox{colorhex1}{\makebox[4.2cm][l]{\geminitwofive\ }}} & 47.81&0.98&0.59&0.84&0.75 \\
        & {\colorbox{colorhex2}{\makebox[4.2cm][l]{\llamathree$^\dagger$ }}} & 46.04&0.98&0.56&0.83&0.74 \\
        & {\colorbox{colorhex2}{\makebox[4.2cm][l]{\gemmathree$^\dagger$ }}} & 48.49&0.97&0.59&0.84&0.75 \\
        & {\colorbox{colorhex3}{\makebox[4.2cm][l]{\madlad$^\dagger$ }}} & 49.93&0.96&0.61&0.82&0.75 \\
        \cdashlinelr{2-7}
        & {\small \salamandratwo$^\dagger$}  & 40.92 & 0.93 & 0.54 & 0.81 & 0.73 \\
        & \quad + {\small \acadtrain\ } & 51.29&1.00& 0.63 &0.83&0.74 \\
        \cdashlinelr{2-7}
        & {\small \salamandraseven$^\dagger$ } & 49.62 & 1.00 & 0.63 & 0.83 & 0.75   \\
        & \quad + {\small \acadtrain\ } &  53.14 & 1.00 & 0.65 & 0.83 & 0.74   \\
\cdashlinelr{1-7}
{fr→en} 
        & {\colorbox{colorhex1}{\makebox[4.2cm][l]{\gptmini\ }}} & 38.76&0.99&0.54&0.83&0.79  \\
        & {\colorbox{colorhex1}{\makebox[4.2cm][l]{\gptnano\ }}} & 35.11&0.97&0.51&0.83&0.79  \\
        & {\colorbox{colorhex1}{\makebox[4.2cm][l]{\geminitwo\ }}}	 & 41.15&1.0&0.56&0.83&0.79 \\
        & {\colorbox{colorhex1}{\makebox[4.2cm][l]{\geminitwofive\ }}} & 37.92&0.97&0.53&0.83&0.79 \\
        & {\colorbox{colorhex2}{\makebox[4.2cm][l]{\llamathree$^\dagger$ }}} & 36.48&1.0&0.51&0.82&0.77 \\
        & {\colorbox{colorhex2}{\makebox[4.2cm][l]{\gemmathree$^\dagger$ }}} & 38.63&0.98&0.54&0.83&0.79 \\
        & {\colorbox{colorhex3}{\makebox[4.2cm][l]{\madlad$^\dagger$ }}} & 36.87&0.93&0.5&0.81&0.79 \\
        \cdashlinelr{2-7}
        & {\small \salamandratwo$^\dagger$ } & 32.30&0.93&0.48&0.81&0.77  \\
        & \quad + {\small \acadtrain\ } & 41.66&0.99&0.56&0.82&0.78 \\
        \cdashlinelr{2-7}
        & {\small \salamandraseven$^\dagger$ } & 38.78&0.98&0.54&0.82&0.78  \\
        & \quad + {\small \acadtrain\ } & 40.61&1.00&0.55&0.82&0.77 \\
\cdashlinelr{1-7}
{ca→en} 
        & {\colorbox{colorhex1}{\makebox[4.2cm][l]{\gptmini\ }}} & 48.85&1.00&0.62&0.85&0.76 \\
        & {\colorbox{colorhex1}{\makebox[4.2cm][l]{\gptnano\ }}} & 43.20&0.97&0.58&0.85&0.77  \\
        & {\colorbox{colorhex1}{\makebox[4.2cm][l]{\geminitwo\ }}}	 & 51.10&1.0&0.64&0.85&0.76  \\
        & {\colorbox{colorhex1}{\makebox[4.2cm][l]{\geminitwofive\ }}} & 47.67&0.98&0.61&0.85&0.77 \\
        & {\colorbox{colorhex2}{\makebox[4.2cm][l]{\llamathree$^\dagger$ }}} & 44.86&1.0&0.58&0.84&0.76 \\
        & {\colorbox{colorhex2}{\makebox[4.2cm][l]{\gemmathree$^\dagger$ }}} & 49.54&0.99&0.62&0.85&0.76 \\
        & {\colorbox{colorhex3}{\makebox[4.2cm][l]{\madlad$^\dagger$ }}} & 32.12&0.74&0.45&0.8&0.77 \\
        \cdashlinelr{2-7}
        & {\small \salamandratwo$^\dagger$ } & 37.21&0.92&0.52&0.83&0.75  \\
        & \quad + {\small \acadtrain\ } & 50.63&1.0&0.63&0.84&0.75 \\
        \cdashlinelr{2-7}
        & {\small \salamandraseven$^\dagger$ } & 47.36&1.0&0.6&0.84&0.75  \\
        & \quad + {\small \acadtrain\ } & 52.42&1.0&0.64&0.85&0.75 \\
\bottomrule
\end{tabular}
}
\caption{ Translation results for the xx$\rightarrow$en language pairs in \acadbench\ dataset. Baselines are grouped into \colorbox{colorhex3}{\textbf{dedicated MMNMT models}}, \colorbox{colorhex2}{\textbf{medium- to small-sized open-weights models}} and \colorbox{colorhex1}{\textbf{large-scale proprietary general models}}. Models with open weights are marked with $^\dagger$.  }
\label{tab:results_xx_en}
\end{table*}

\begin{table*}[!ht]
\centering
\small
\tabcolsep 13pt
\scalebox{0.97}{%
\begin{tabular}{llrrrr}
\toprule
Pair & Model & {\small d-}\textsc{BLEU} & \textsc{BP} & \textsc{Comet} & \textsc{Comet-}\textsc{\scriptsize Kiwi} \\
\midrule
{en→es} 
        & {\colorbox{colorhex1}{\makebox[4.2cm][l]{\gptmini\ }}} & 50.60 	& 0.98 & 0.86 & 0.83  \\
        & {\colorbox{colorhex1}{\makebox[4.2cm][l]{\gptnano\ }}} &  48.83	& 0.99 & 0.86 & 0.82  \\
        & {\colorbox{colorhex1}{\makebox[4.2cm][l]{\geminitwo\ }}}	 & 52.48&0.99&0.86&0.82 \\
        & {\colorbox{colorhex1}{\makebox[4.2cm][l]{\geminitwofive\ }}} & 52.02&0.98&0.87&0.82 \\
        & {\colorbox{colorhex2}{\makebox[4.2cm][l]{\llamathree$^\dagger$ }}} & 44.76&0.98&0.85&0.81 \\
        & {\colorbox{colorhex2}{\makebox[4.2cm][l]{\gemmathree$^\dagger$ }}} & 50.74&0.98&0.86&0.82 \\
        & {\colorbox{colorhex3}{\makebox[4.2cm][l]{\madlad$^\dagger$ }}} & 34.88&0.73&0.82&0.8 \\
        \cdashlinelr{2-6}
        & {\small \salamandratwo$^\dagger$ } & 34.47	& 0.84 & 0.83 & 0.79 \\
        & \quad + {\small \acadtrain\ } & 51.31 & 0.97 & 0.86 & 0.81 \\
        \cdashlinelr{2-6}
        & {\small \salamandraseven$^\dagger$ } & 46.42 & 0.96 & 0.86 & 0.81  \\
        & \quad + {\small \acadtrain\ } & 53.47 & 0.97 & 0.86 & 0.81 \\
\cdashlinelr{1-6}
{en→pt}  & {\colorbox{colorhex1}{\makebox[4.2cm][l]{\gptmini\ }}} &  47.38&1.0&0.86&0.81\\
        & {\colorbox{colorhex1}{\makebox[4.2cm][l]{\gptnano\ }}} & 45.23&1.0&0.86&0.82 \\
        & {\colorbox{colorhex1}{\makebox[4.2cm][l]{\geminitwo\ }}}	 & 48.18&1.0&0.87&0.81  \\
        & {\colorbox{colorhex1}{\makebox[4.2cm][l]{\geminitwofive\ }}} & 48.29&1.0&0.86&0.81 \\
        & {\colorbox{colorhex2}{\makebox[4.2cm][l]{\llamathree$^\dagger$ }}} & 41.81&0.99&0.85&0.8 \\
        & {\colorbox{colorhex2}{\makebox[4.2cm][l]{\gemmathree$^\dagger$ }}} & 46.96&1.0&0.86&0.81 \\
        & {\colorbox{colorhex3}{\makebox[4.2cm][l]{\madlad$^\dagger$ }}} & 49.58&0.97&0.86&0.8
 \\
        \cdashlinelr{2-6}
        & {\small \salamandratwo$^\dagger$ }  & 37.12&0.96 &0.84&0.77 \\
        & \quad + {\small \acadtrain\ } & 48.18&0.99 &0.86&0.81 \\
        \cdashlinelr{2-6}
        & {\small \salamandraseven$^\dagger$ } & 43.73&1.0&0.85&0.81 \\
        & \quad + {\small \acadtrain\ } & 50.26&0.99&0.86&0.81  \\
\cdashlinelr{1-6}
{en→fr} 
        & {\colorbox{colorhex1}{\makebox[4.2cm][l]{\gptmini\ }}} &  37.55&1.0&0.86&0.84 \\
        & {\colorbox{colorhex1}{\makebox[4.2cm][l]{\gptnano\ }}} &  37.05&1.0&0.85&0.83 \\
        & {\colorbox{colorhex1}{\makebox[4.2cm][l]{\geminitwo\ }}}	& 42.36&0.99&0.86&0.83  \\
        & {\colorbox{colorhex1}{\makebox[4.2cm][l]{\geminitwofive\ }}} & 41.58&0.99&0.86&0.83 \\
        & {\colorbox{colorhex2}{\makebox[4.2cm][l]{\llamathree$^\dagger$ }}} & 36.14&0.99&0.84&0.82 \\
        & {\colorbox{colorhex2}{\makebox[4.2cm][l]{\gemmathree$^\dagger$ }}} & 40.05&1.0&0.86&0.83 \\
        & {\colorbox{colorhex3}{\makebox[4.2cm][l]{\madlad$^\dagger$ }}} & 37.3&0.92&0.84&0.82 \\
        \cdashlinelr{2-6}
        & {\small \salamandratwo$^\dagger$ } & 29.73&0.93&0.81&0.79  \\
        & \quad + {\small \acadtrain\ } & 40.05&0.97&0.84&0.82  \\
        \cdashlinelr{2-6}
        & {\small \salamandraseven$^\dagger$ } & 37.19&0.97&0.84&0.82 \\
        & \quad + {\small \acadtrain\ } & 42.28&0.96&0.85&0.82 \\
\cdashlinelr{1-6}
{en→ca} 
        & {\colorbox{colorhex1}{\makebox[4.2cm][l]{\gptmini\ }}} & 44.51&0.98&0.88&0.82 \\
        & {\colorbox{colorhex1}{\makebox[4.2cm][l]{\gptnano\ }}} & 44.03&0.99&0.88&0.82  \\
        & {\colorbox{colorhex1}{\makebox[4.2cm][l]{\geminitwo\ }}}	& 48.99&0.99&0.88&0.82  \\
        & {\colorbox{colorhex1}{\makebox[4.2cm][l]{\geminitwofive\ }}} & 49.12&0.99&0.88&0.82 \\
        & {\colorbox{colorhex2}{\makebox[4.2cm][l]{\llamathree$^\dagger$ }}} & 36.76&1.0&0.86&0.8 \\
        & {\colorbox{colorhex2}{\makebox[4.2cm][l]{\gemmathree$^\dagger$ }}} & 47.41&0.99&0.88&0.82 \\
        & {\colorbox{colorhex3}{\makebox[4.2cm][l]{\madlad$^\dagger$ }}} & 22.54&0.65&0.79&0.77  \\
        \cdashlinelr{2-6}
        & {\small \salamandratwo$^\dagger$ } & 30.33&0.87&0.85&0.78 \\
        & \quad + {\small \acadtrain\ } & 47.90&0.97&0.88&0.81 \\
        \cdashlinelr{2-6}
        & {\small \salamandraseven$^\dagger$ }  & 42.87&0.97&0.87&0.81 \\
        & \quad + {\small \acadtrain\ } & 50.77&0.98&0.88&0.81 \\
\bottomrule
\end{tabular}
}
\caption{  Translation results for the en$\rightarrow$xx language pairs in \acadbench\ dataset. Baselines are grouped into \colorbox{colorhex3}{\textbf{dedicated MMNMT models}}, \colorbox{colorhex2}{\textbf{medium- to small-sized open-weights models}} and \colorbox{colorhex1}{\textbf{large-scale proprietary general models}}. Models with open weights are marked with $^\dagger$.  }
\label{tab:results_en_xx}
\end{table*}

\begin{table*}[!ht]
\centering
\small
\tabcolsep 13pt
\begin{tabular}{llrrrr}
\toprule
Pair & Model & {\small d-}\textsc{BLEU} & \textsc{BP} & \textsc{Comet} & \textsc{Comet-}\textsc{\scriptsize Kiwi} \\
\midrule
{fr→es} 
        & {\colorbox{colorhex1}{\makebox[4.2cm][l]{\gptmini\ }}} & 48.54&0.97&0.84&0.84 \\
        & {\colorbox{colorhex1}{\makebox[4.2cm][l]{\gptnano\ }}} & 46.74&0.97&0.84&0.84  \\
        & {\colorbox{colorhex1}{\makebox[4.2cm][l]{\geminitwo\ }}} & 48.78&0.98&0.84&0.84  \\
        & {\colorbox{colorhex1}{\makebox[4.2cm][l]{\geminitwofive\ }}} & 48.97&0.97&0.84&0.84 \\
        & {\colorbox{colorhex2}{\makebox[4.2cm][l]{\llamathree$^\dagger$ }}} & 44.26&0.97&0.83&0.83 \\
        & {\colorbox{colorhex2}{\makebox[4.2cm][l]{\gemmathree$^\dagger$ }}} & 48.66&0.97&0.84&0.84 \\
        & {\colorbox{colorhex3}{\makebox[4.2cm][l]{\madlad$^\dagger$ }}} & 50.84&0.98&0.84&0.84 \\
        \cdashlinelr{2-6}
        & {\small \salamandratwo$^\dagger$ } & 41.68&0.95&0.83&0.83 \\
        & \quad + {\small \acadtrain\ } & 49.69&0.97&0.84&0.84 \\
        \cdashlinelr{2-6}
        & {\small \salamandraseven$^\dagger$ } & 45.76&0.97&0.84&0.84  \\
        & \quad + {\small \acadtrain\ } & 51.51&0.96&0.84&0.84 \\
\cdashlinelr{1-6}
{ca→es} 
        & {\colorbox{colorhex1}{\makebox[4.2cm][l]{\gptmini\ }}} & 82.93&1.0&0.89&0.8 \\
        & {\colorbox{colorhex1}{\makebox[4.2cm][l]{\gptnano\ }}} & 78.07&1.0&0.89&0.8  \\
        & {\colorbox{colorhex1}{\makebox[4.2cm][l]{\geminitwo\ }}}	 & 84.8&1.0&0.89&0.8  \\
        & {\colorbox{colorhex1}{\makebox[4.2cm][l]{\geminitwofive\ }}} & 83.3&1.0&0.89&0.8 \\
        & {\colorbox{colorhex2}{\makebox[4.2cm][l]{\llamathree$^\dagger$ }}} & 77.19&1.0&0.89&0.8 \\
        & {\colorbox{colorhex2}{\makebox[4.2cm][l]{\gemmathree$^\dagger$ }}} & 82.74&1.0&0.89&0.8 \\
        & {\colorbox{colorhex3}{\makebox[4.2cm][l]{\madlad$^\dagger$ }}} & 44.6&0.58&0.84&0.78 \\
        \cdashlinelr{2-6}
        & {\small \salamandratwo$^\dagger$ } &  74.12&0.96&0.88&0.79 \\
        & \quad + {\small \acadtrain\ } & 84.9&1.0&0.89&0.8 \\
        \cdashlinelr{2-6}
        & {\small \salamandraseven$^\dagger$ }  & 80.48&1.0&0.89&0.8 \\
        & \quad + {\small \acadtrain\ } & 85.83&1.0&0.89&0.8 \\
\bottomrule
\end{tabular}
\caption{Translation results for the xx$\rightarrow$es language pairs in \acadbench\ dataset. Baselines are grouped into \colorbox{colorhex3}{\textbf{dedicated MMNMT models}}, \colorbox{colorhex2}{\textbf{medium- to small-sized open-weights models}} and \colorbox{colorhex1}{\textbf{large-scale proprietary general models}}. Models with open weights are marked with $^\dagger$.   }
\label{tab:results_xx_es}
\end{table*}

\begin{table*}[!ht]
\centering
\small
\tabcolsep 13pt
\begin{tabular}{llrrrr}
\toprule
Pair & Model & {\small d-}\textsc{BLEU} & \textsc{BP} & \textsc{Comet} & \textsc{Comet-}\textsc{\scriptsize Kiwi} \\
\midrule
{es→fr} 
        & {\colorbox{colorhex1}{\makebox[4.2cm][l]{\gptmini\ }}} & 43.10&0.99&0.84&0.84 \\
        & {\colorbox{colorhex1}{\makebox[4.2cm][l]{\gptnano\ }}} & 42.42&1.0&0.84&0.84  \\
        & {\colorbox{colorhex1}{\makebox[4.2cm][l]{\geminitwo\ }}} & 48.27&0.99&0.84&0.84 \\
        & {\colorbox{colorhex1}{\makebox[4.2cm][l]{\geminitwofive\ }}} & 46.39&0.99&0.84&0.84 \\
        & {\colorbox{colorhex2}{\makebox[4.2cm][l]{\llamathree$^\dagger$ }}} & 41.91&1.0&0.83&0.83 \\
        & {\colorbox{colorhex2}{\makebox[4.2cm][l]{\gemmathree$^\dagger$ }}} & 44.08&0.98&0.84&0.84 \\
        & {\colorbox{colorhex3}{\makebox[4.2cm][l]{\madlad$^\dagger$ }}} & 45.56&1.0&0.84&0.84 \\
        \cdashlinelr{2-6}
        & {\small \salamandratwo$^\dagger$ } & 40.16&0.97&0.82&0.82 \\
        & \quad + {\small \acadtrain\ } & 48.57&0.99&0.84&0.84 \\
        \cdashlinelr{2-6}
        & {\small \salamandraseven$^\dagger$ }  & 44.3&0.98&0.84&0.84  \\
        & \quad + {\small \acadtrain\ } &  50.2&0.99&0.84&0.84 \\
\cdashlinelr{1-6}
{es→ca} 
        & {\colorbox{colorhex1}{\makebox[4.2cm][l]{\gptmini\ }}} & 71.4&0.99&0.9&0.82  \\
        & {\colorbox{colorhex1}{\makebox[4.2cm][l]{\gptnano\ }}} & 70.61&0.99&0.9&0.82 \\
        & {\colorbox{colorhex1}{\makebox[4.2cm][l]{\geminitwo\ }}}	 & 81.29&0.99&0.9&0.82  \\
        & {\colorbox{colorhex1}{\makebox[4.2cm][l]{\geminitwofive\ }}} & 79.46&0.99&0.9&0.82 \\
        & {\colorbox{colorhex2}{\makebox[4.2cm][l]{\llamathree$^\dagger$ }}} & 69.34&0.99&0.89&0.81 \\
        & {\colorbox{colorhex2}{\makebox[4.2cm][l]{\gemmathree$^\dagger$ }}} & 79.03&0.99&0.9&0.82 \\
        & {\colorbox{colorhex3}{\makebox[4.2cm][l]{\madlad$^\dagger$ }}} & 38.99&0.57&0.84&0.81 \\
        \cdashlinelr{2-6}
        & {\small \salamandratwo$^\dagger$ } &  65.43&0.92&0.88&0.81 \\
        & \quad + {\small \acadtrain\ } & 80.42&0.98&0.9&0.82 \\
        \cdashlinelr{2-6}
        & {\small \salamandraseven$^\dagger$ } & 74.93&0.98&0.9&0.81 \\
        & \quad + {\small \acadtrain\ } & 81.58&0.99&0.9&0.82 \\
\bottomrule
\end{tabular}
\caption{ Translation results for the es$\rightarrow$xx language pairs in \acadbench\ dataset. Baselines are grouped into \colorbox{colorhex3}{\textbf{dedicated MMNMT models}}, \colorbox{colorhex2}{\textbf{medium- to small-sized open-weights models}} and \colorbox{colorhex1}{\textbf{large-scale proprietary general models}}. Models with open weights are marked with $^\dagger$.   }
\label{tab:results_es_xx}
\end{table*}

\subsection{Results \acadtrain\ directions}

Tables \ref{tab:wmt_bleu_v1} and \ref{tab:wmt_bleu_v2} present the \textsc{BLEU} scores, and Tables \ref{tab:wmt_comet_v1} and \ref{tab:wmt_comet_v2} report the \textsc{COMET} scores for the translation directions common to both \wmtplusplus\ and \acadtrain.

\begin{sidewaystable}
\begin{center}
\tiny
\tabcolsep 5pt
\begin{tabular}{l*{31}{c}}
\textbf{Model} &
\rotatebox{-90}{en-es} & \rotatebox{-90}{en-fr} & \rotatebox{-90}{en-ca} & \rotatebox{-90}{es-ca}  & \rotatebox{-90}{es-fr} & \rotatebox{-90}{en-pt} & \rotatebox{-90}{es-pt} & \rotatebox{-90}{de-fr} &  
\rotatebox{-90}{en-de} & \rotatebox{-90}{en-it} & \rotatebox{-90}{fr-pt} & \rotatebox{-90}{fr-it} & \rotatebox{-90}{es-de} & \rotatebox{-90}{ca-fr} & \rotatebox{-90}{es-it} & \rotatebox{-90}{de-it} &
\rotatebox{-90}{fr-nl} & \rotatebox{-90}{fr-el} & \rotatebox{-90}{en-el} & \rotatebox{-90}{nl-en} & \rotatebox{-90}{it-pt} & \rotatebox{-90}{de-pt} & \rotatebox{-90}{ca-pt} & \rotatebox{-90}{ca-it} & \rotatebox{-90}{ca-de}  & \rotatebox{-90}{es-nl} & \rotatebox{-90}{it-nl} & \rotatebox{-90}{el-es} &
\rotatebox{-90}{nl-de} & \rotatebox{-90}{ca-el} & \rotatebox{-90}{ca-nl} \\
\midrule
{\salamandratwo{}}  & 30.0& 25.7& 24.6& 32.6& 23.6& 21.9& 22.2& 18.0& 19.0& 25.4& 19.4& 22.7& 15.6& 22.4& 24.3& 16.6& 17.7& 14.4& 21.2& 28.3& 21.4& 15.4& 21.8& 14.7& 15.5& 17.5& 17.6& 23.7& 17.9& 16.4& 17.1 \\ \\
\quad + {\acadtrain\ } & 31.0& 26.7& 28.8& 34.5& 26.9& 23.9& 24.8& 22.5& 16.2& 25.0& 23.4& 25.5& 17.4& 26.7& 28.2& 20.6& 18.6& 16.2& 20.4& 30.0& 25.5& 19.8& 26.0& 27.7& 17.7& 19.0& 19.3& 27.3& 21.0& 17.6& 18.9 \\ \\
\cdashlinelr{1-32} \\
{\salamandraseven{} } & 38.6& 33.2& 35.3& 36.7& 29.0& 26.3& 24.6& 24.7& 26.1& 34.5& 22.0& 28.0& 23.1& 28.3& 30.1& 23.6& 24.1& 21.9& 29.7& 33.3& 24.5& 19.4& 25.4& 29.1& 21.4& 24.5& 24.4& 29.8& 23.0& 23.4& 24.0 \\ \\
\quad + {\acadtrain\ } & 40.2& 35.5& 37.9& 36.5& 31.7& 29.8& 27.3& 26.8& 26.0& 35.0& 25.0& 28.2& 22.4& 29.8& 32.5& 24.0& 23.7& 23.4& 30.2& 33.9& 27.1& 22.5& 27.7& 30.6& 21.7& 25.3& 24.1& 32.0& 24.0& 25.6& 23.7 \\
\bottomrule
\end{tabular}

\end{center}
\caption{\bleu\ results on \wmtplusplus\ on all the directions shared between \wmtplusplus\ and \acadtrain.}
\label{tab:wmt_bleu_v1}
\end{sidewaystable}

\begin{sidewaystable}
\begin{center}
\tiny
\tabcolsep 5pt
\begin{tabular}{l*{31}{c}}
\textbf{Model} &
\rotatebox{-90}{es-en} & \rotatebox{-90}{fr-en} & \rotatebox{-90}{ca-en} & \rotatebox{-90}{ca-es} & \rotatebox{-90}{fr-es} & \rotatebox{-90}{pt-en} & \rotatebox{-90}{pt-es} & \rotatebox{-90}{fr-de} &
\rotatebox{-90}{de-en} & \rotatebox{-90}{it-en} & \rotatebox{-90}{pt-fr} & \rotatebox{-90}{it-fr} & \rotatebox{-90}{de-es} & \rotatebox{-90}{fr-ca} & \rotatebox{-90}{it-es} & \rotatebox{-90}{it-de} & 
\rotatebox{-90}{nl-fr} & \rotatebox{-90}{el-fr} & \rotatebox{-90}{el-en} & \rotatebox{-90}{en-nl} & \rotatebox{-90}{pt-it} & \rotatebox{-90}{pt-de} & \rotatebox{-90}{pt-ca} & \rotatebox{-90}{it-ca} & \rotatebox{-90}{de-ca} & \rotatebox{-90}{nl-es} & \rotatebox{-90}{nl-it} & \rotatebox{-90}{es-el} &
\rotatebox{-90}{de-nl} & \rotatebox{-90}{el-ca}  & \rotatebox{-90}{nl-ca} \\
\midrule
{\salamandratwo{}}  & 34.9& 29.3& 31.0& 33.2& 24.4& 29.9& 27.2& 14.5& 26.5& 31.4& 22.8& 24.3& 19.7& 23.1& 27.1& 14.5& 18.8& 19.9& 31.5& 21.9& 22.5& 15.4& 26.0& 25.2& 19.0& 20.2& 16.9& 16.7& 19.4& 22.3& 17.6 \\ \\
\quad + {\acadtrain\ } & 35.2& 31.6& 34.3& 36.5& 26.2& 31.4& 29.1& 17.4& 29.0& 35.8& 26.3& 28.1& 20.2& 26.4& 30.9& 19.0& 21.6& 24.3& 35.5& 20.0& 25.9& 17.1& 29.0& 29.2& 21.1& 20.7& 20.8& 17.6& 20.0& 25.7& 22.5 \\ \\
\cdashlinelr{1-32} \\
{\salamandraseven{} } & 41.5& 34.7& 37.6& 36.1& 28.7& 34.8& 29.7& 20.7& 31.8& 39.1& 28.3& 29.8& 25.2& 28.3& 31.8& 21.8& 25.6& 27.2& 40.4& 29.0& 28.4& 20.7& 30.9& 31.0& 25.1& 26.1& 24.2& 23.9& 24.1& 29.1& 26.2 \\ \\
\quad + {\acadtrain\ } & 41.9& 35.1& 37.3& 37.9& 29.9& 35.0& 31.0& 22.2& 32.0& 39.3& 29.7& 30.7& 26.9& 28.9& 33.3& 23.2& 27.4& 29.3& 40.6& 26.0& 28.9& 22.5& 31.8& 31.7& 26.5& 27.9& 24.6& 24.4& 23.3& 31.0& 27.8 \\
\bottomrule
\end{tabular}
\end{center}
\caption{\bleu\ results on \wmtplusplus\ for the reverse translation directions corresponding to those shown in Table~\ref{tab:wmt_bleu_v1}. }
\label{tab:wmt_bleu_v2}
\end{sidewaystable}

\begin{sidewaystable}
\begin{center}
\tiny
\tabcolsep 5pt
\begin{tabular}{l*{31}{c}}
\textbf{Model} &
\rotatebox{-90}{en-es} & \rotatebox{-90}{en-fr} & \rotatebox{-90}{en-ca} & \rotatebox{-90}{es-ca}  & \rotatebox{-90}{es-fr} & \rotatebox{-90}{en-pt} & \rotatebox{-90}{es-pt} & \rotatebox{-90}{de-fr} &  
\rotatebox{-90}{en-de} & \rotatebox{-90}{en-it} & \rotatebox{-90}{fr-pt} & \rotatebox{-90}{fr-it} & \rotatebox{-90}{es-de} & \rotatebox{-90}{ca-fr} & \rotatebox{-90}{es-it} & \rotatebox{-90}{de-it} &
\rotatebox{-90}{fr-nl} & \rotatebox{-90}{fr-el} & \rotatebox{-90}{en-el} & \rotatebox{-90}{nl-en} & \rotatebox{-90}{it-pt} & \rotatebox{-90}{de-pt} & \rotatebox{-90}{ca-pt} & \rotatebox{-90}{ca-it} & \rotatebox{-90}{ca-de}  & \rotatebox{-90}{es-nl} & \rotatebox{-90}{it-nl} & \rotatebox{-90}{el-es} &
\rotatebox{-90}{nl-de} & \rotatebox{-90}{ca-el} & \rotatebox{-90}{ca-nl} \\
\midrule
{\salamandratwo{}} & 0.78& 0.74& 0.76& 0.81& 0.75& 0.77& 0.79& 0.73& 0.72& 0.77& 0.77& 0.79& 0.72& 0.73& 0.80& 0.76& 0.75& 0.71& 0.75& 0.81& 0.78& 0.76& 0.78& 0.76& 0.71& 0.75& 0.75& 0.78& 0.74& 0.73& 0.73 \\ \\
\quad + {\acadtrain\ } & 0.77& 0.74& 0.75& 0.82& 0.77& 0.76& 0.81& 0.75& 0.71& 0.75& 0.79& 0.80& 0.75& 0.76& 0.81& 0.78& 0.76& 0.75& 0.72& 0.82& 0.80& 0.77& 0.79& 0.80& 0.73& 0.76& 0.77& 0.79& 0.78& 0.74& 0.74 \\ \\
\cdashlinelr{1-32} \\
{\salamandraseven{} } & 0.81& 0.79& 0.80& 0.83& 0.79& 0.80& 0.82& 0.77& 0.79& 0.81& 0.80& 0.82& 0.78& 0.78& 0.82& 0.80& 0.80& 0.80& 0.81& 0.84& 0.81& 0.79& 0.81& 0.82& 0.77& 0.80& 0.81& 0.81& 0.80& 0.81& 0.79 \\ \\
\quad + {\acadtrain\ } & 0.84& 0.83& 0.83& 0.83& 0.80& 0.83& 0.82& 0.78& 0.83& 0.84& 0.79& 0.80& 0.79& 0.80& 0.82& 0.78& 0.78& 0.78& 0.84& 0.78& 0.82& 0.79& 0.82& 0.82& 0.78& 0.80& 0.80& 0.81& 0.82& 0.80& 0.79 \\
\bottomrule
\end{tabular}
\end{center}
\caption{\comet\ results on \wmtplusplus\ on all the directions shared between \wmtplusplus\ and \acadtrain.}
\label{tab:wmt_comet_v1}
\end{sidewaystable}

\begin{sidewaystable}
\begin{center}
\tiny
\tabcolsep 5pt
\begin{tabular}{l*{31}{c}}
\textbf{Model} &
\rotatebox{-90}{es-en} & \rotatebox{-90}{fr-en} & \rotatebox{-90}{ca-en} & \rotatebox{-90}{ca-es} & \rotatebox{-90}{fr-es} & \rotatebox{-90}{pt-en} & \rotatebox{-90}{pt-es} & \rotatebox{-90}{fr-de} &
\rotatebox{-90}{de-en} & \rotatebox{-90}{it-en} & \rotatebox{-90}{pt-fr} & \rotatebox{-90}{it-fr} & \rotatebox{-90}{de-es} & \rotatebox{-90}{fr-ca} & \rotatebox{-90}{it-es} & \rotatebox{-90}{it-de} & 
\rotatebox{-90}{nl-fr} & \rotatebox{-90}{el-fr} & \rotatebox{-90}{el-en} & \rotatebox{-90}{en-nl} & \rotatebox{-90}{pt-it} & \rotatebox{-90}{pt-de} & \rotatebox{-90}{pt-ca} & \rotatebox{-90}{it-ca} & \rotatebox{-90}{de-ca} & \rotatebox{-90}{nl-es} & \rotatebox{-90}{nl-it} & \rotatebox{-90}{es-el} &
\rotatebox{-90}{de-nl} & \rotatebox{-90}{el-ca}  & \rotatebox{-90}{nl-ca} \\
\midrule
{\salamandratwo{}}  & 0.82& 0.81& 0.80& 0.81& 0.78& 0.81& 0.81& 0.72& 0.81& 0.81& 0.75& 0.76& 0.76& 0.76& 0.79& 0.72& 0.73& 0.74& 0.82& 0.77& 0.80& 0.73& 0.79& 0.78& 0.74& 0.76& 0.75& 0.73& 0.79& 0.77& 0.74 \\ \\
\quad + {\acadtrain\ } & 0.83& 0.81& 0.81& 0.82& 0.79& 0.82& 0.82& 0.75& 0.82& 0.83& 0.77& 0.78& 0.77& 0.78& 0.81& 0.75& 0.75& 0.76& 0.83& 0.74& 0.81& 0.74& 0.80& 0.80& 0.76& 0.77& 0.78& 0.76& 0.80& 0.78& 0.76  \\ \\
\cdashlinelr{1-32} \\
{\salamandraseven{} } & 0.85& 0.83& 0.83& 0.83& 0.81& 0.84& 0.82& 0.78& 0.83& 0.84& 0.79& 0.80& 0.79& 0.80& 0.82& 0.79& 0.77& 0.78& 0.85& 0.81& 0.82& 0.78& 0.81& 0.81& 0.78& 0.80& 0.80& 0.81& 0.83& 0.80& 0.79 \\ \\
\quad + {\acadtrain\ } & 0.84& 0.83& 0.83& 0.83& 0.80& 0.83& 0.82& 0.78& 0.83& 0.84& 0.79& 0.80& 0.79& 0.80& 0.82& 0.78& 0.78& 0.78& 0.84& 0.78& 0.82& 0.79& 0.82& 0.82& 0.78& 0.80& 0.80& 0.81& 0.82& 0.80& 0.79  \\
\bottomrule
\end{tabular}
\end{center}
\caption{\comet\ results on \wmtplusplus\ for the reverse translation directions corresponding to those shown in Table~\ref{tab:wmt_comet_v1}.}
\label{tab:wmt_comet_v2}
\end{sidewaystable}


\end{document}